\documentclass[a4paper, fleqn]{cas-sc}
\usepackage{amsmath} 
\makeatother
\usepackage[normalem]{ulem}
\useunder{\uline}{\ul}{}
\usepackage{indentfirst}
\usepackage{geometry}
\usepackage{lipsum} 
\usepackage{lineno}
\usepackage{tabularx}
\usepackage{booktabs}
\usepackage{float}
\usepackage{CJKutf8}

\usepackage[utf8]{inputenc}

\usepackage[authoryear]{natbib}
\usepackage{amsmath}

\usepackage{autobreak}
\usepackage{amsfonts}
\usepackage{diagbox}
\usepackage{multirow}
\usepackage{makecell}
\usepackage{array}
\usepackage{bm}
\usepackage{graphicx}
\usepackage{algorithm}
\usepackage{algorithmic}
\usepackage{caption}
\usepackage{subcaption}
\usepackage{pdflscape}
\usepackage{enumitem}
\usepackage{hyperref}
\usepackage{cleveref}
\crefname{table}{Table}{Table.}
\crefname{figure}{Fig.}{Fig.}
\crefname{subfigure}{Fig.}{Fig.}
\crefname{equation}{Eq.}{Eq.}
\crefname{section}{Section}{Sect.}
\crefname{appendix}{Appendix}{App.}
\crefname{subsection}{Subsection}{subsect.}
\crefname{algorithm}{Algorithm.}{Algorithm.}
\newcommand{\perfup}[2]{#1 \small \newline({+#2\%})}
\newcommand{\perfdown}[2]{#1 \small \newline({-#2\%})}
\newcommand{\perfsame}[2]{#1 \small \newline({#2\%})}
\newcolumntype{Y}{>{\centering\arraybackslash}X}

\newcommand{\nop}[1]{}

\def\tsc#1{\csdef{#1}{\textsc{\lowercase{#1}}\xspace}}
\tsc{WGM}
\tsc{QE}



\begin{document}
\let\WriteBookmarks\relax
\renewcommand{\floatpagefraction}{0.8}
\renewcommand{\textfraction}{0.1}
\renewcommand{\topfraction}{0.9}
\renewcommand{\bottomfraction}{0.8}

\shorttitle{Resilient and Trustworthy Traffic Flow Inference}

\shortauthors{Zhou et al.}  

\title[mode = title]{Metropolis-Scale Resilient and Trustworthy Traffic Flow Inference Using Multi-Source Data}


\author[1]{Qishen Zhou}[auid=000,orcid=0000-0001-8074-4111]
\ead{qishenzhou@jlu.edu.cn}
\credit{Conceptualization, Methodology, Software, Investigation, Formal analysis, Writing – original draft}

\author[2]{Yifan Zhang}[auid=001, orcid=0000-0001-8882-4327]
\ead{yifan.zhang@cityu-dg.edu.cn}
\credit{Conceptualization, Methodology, Writing – review \& editing}

\author[3]{Michail A. Makridis}[auid=002, orcid=0000-0001-7462-4674]
\ead{mmakridis@ethz.ch}
\credit{Supervision, Writing – review \& editing}

\author[3]{Anastasios Kouvelas}[auid=003, orcid=0000-0003-4571-2530]
\ead{kouvelas@ethz.ch}
\credit{Supervision, Writing – review \& editing}

\author[4]{Yibing Wang}[auid=004]
\ead{wangyibing@zju.edu.cn}
\credit{Supervision, Writing – review \& editing}

\author[5]{Simon Hu}[auid=005,orcid=0000-0002-9832-6679]
\cormark[1]
\ead{simonhu@zju.edu.cn}
\credit{Funding acquisition, Resources, Supervision, Writing – review \& editing}


\affiliation[a]{organization={School of Transportation, Jilin University},
            city={Changchun},
            postcode={130022},
            country={China}}

\affiliation[b]{organization={Department of Computer Science, City University of Hong Kong (Dongguan)},
            city={Dongguan},
            postcode={523808},
            country={China}}

\affiliation[c]{organization={Institute for Transport Planning and Systems, ETH Zurich},
            city={Zurich},
            postcode={8093}, 
            country={Switzerland}}
            
\affiliation[d]{organization={Institute of Intelligent Transportation Systems, College of Civil Engineering and Architecture, Zhejiang University},
            city={Hangzhou},
            postcode={310058}, 
            country={China}}    
            
\affiliation[e]{organization={ZJU-UIUC Institute, Zhejiang University},
            city={Haining},
            postcode={314400},
            country={China}}

\cortext[1]{Corresponding author}



\begin{abstract}
Inferring network-wide traffic states from sparse observations with high accuracy and trustworthy uncertainty quantification is essential for intelligent transportation systems, yet it remains challenging due to the underdetermined nature of the problem, multifaceted disturbances in sensing networks, and the inherent conflicts among multiple inference subtasks when modeled jointly. We propose the Task-Aware Attentive Neural Process (TA-ANP), a unified probabilistic framework for resilient and trustworthy global traffic state inference (GTSI) by fusing floating car data (FCD) with sparse fixed-detector measurements. By casting GTSI as a stochastic process, TA-ANP leverages the meta-learning properties of neural processes to adapt rapidly to changes in sensing configurations without retraining. A task-aware multi-query attention module with distinct spatiotemporal inductive biases is introduced to jointly handle three GTSI subtasks---real-time estimation at unobserved locations, forecasting at both observed and unobserved locations---while mitigating cross-task interference. For uncertainty quantification, we combine neural processes with Monte Carlo Dropout to capture both aleatoric and epistemic uncertainty. To support metropolis-scale evaluation, we construct the Metropolitan Multi-Source Traffic Dataset (MMTD), integrating fixed-loop sensor measurements, FCD statistics, and OpenStreetMap road-network data over an urban network of 26,166 road segments, including 2,371 fixed-sensor-equipped segments used for quantitative evaluation. Experiments across multiple independently generated sensor configurations show that TA-ANP achieves best performance across all subtasks under deterministic and probabilistic metrics. The resulting well-calibrated uncertainties enable more efficient fixed-sensor placement with fewer sensor deployments. Under a Damage--Repair--Addition sensing lifecycle, TA-ANP demonstrates superior resilience in terms of disturbance absorption, performance recovery, and adaptability to unseen sensing configurations. The code and processed datasets are publicly available on \href{https://github.com/qishenzhou/TA-ANP}{GitHub} and \href{https://zenodo.org/records/18781181}{Zenodo}, respectively.
\end{abstract}

\begin{highlights}

    \item A neural process-based framework for network-wide traffic state inference.
    \item Tailored attention biases alleviate the seesaw effect across joint traffic inference subtasks.
    \item Aleatoric and epistemic uncertainties are jointly quantified and well-calibrated.
    \item Strong resilience under multifaceted sensing disturbances.
    \item A metropolis-scale multi-source dataset is constructed to benchmark the proposed method.
\end{highlights}


\maketitle
\section{Introduction} \label{Sec:Intro}
Accurate, real-time, and predictive traffic state information is fundamental to network-wide monitoring and control in intelligent transportation systems (ITS). However, the high costs of procuring, installing, and maintaining fixed detectors typically limit deployments to a sparse subset of the network, leaving substantial spatial blind spots. In this context, global traffic state inference (GTSI) plays a pivotal role by inferring current and future traffic states over the entire road network, including unobserved segments, from sparse observations.

In recent years, deep learning has made significant advances in modeling traffic-flow dynamics. Particularly, recurrent neural networks~\citep{R1-1}, graph neural networks~\citep{yu2018STGCN, li2018dcrnn, Wu2020MTGNN, pems37, Zhao2020TGCN, Wang2022STGCN,liuSAFERpredictorSparseAdversarial2025} and transformer-style learners~\citep{DO2019STANN, Liu2023STAEFormer,liu2024itransformer,zhang2025CASAformer} have markedly improved forecast  accuracy  at observed locations. Beyond forecasting, the characterization of observed traffic states has also been actively studied~\citep{R1-2}. Moreover, research is shifting from prediction and state characterization at observed sensors to global state estimation~\citep{liu2019TailoredMachineLearning, xu2023AGNPNetworkwideShortterm, Mei2023UIGNN, zhou2025NetworkWidea, xu2025kits} and forecasting~\citep{YANG2023EKFLSTM,yang2024EKFTLSTM,zhou2025MoGERNNInductiveTraffic,huGlobalOpenMultisource2025}. Despite these advances, several challenges remain.

First, real-world sensing networks are subject to multifaceted disturbances. In the short term, communication outages and environmental interference cause temporary data loss. In the long term, detectors may fail permanently and new sensors could be deployed, leading to structural changes in the sensing network. In practice, these disturbances do not occur in isolation but coexist and evolve continuously, posing a fundamental challenge to the resilience of traffic sensing systems. Although resilience has been extensively studied for transportation infrastructure under natural disasters and sudden incidents,  with emphasis on capacity degradation and post-disruption recovery~\citep{yao2025Timeevolvingb,li2025Adaptive}, much less attention has been paid to the cyber-physical resilience of ITS, the ability of the state-inference layer to remain dependable and functional amidst physical disruptions to fixed sensing infrastructure~\citep{liu2026Quantifying}. This gap is particularly pronounced for traffic sensing networks operating under such multifaceted and evolving disturbances. From a modeling perspective, most existing work targets the imputation of transient missing data~\citep{liang_memory-augmented_2022}, whereas the structural changes of sensing networks remains underexplored~\citep{zhou2025MoGERNNInductiveTraffic}. Furthermore, many existing approaches are transductive~\citep{Bahadori2014KrigingGLTL, zhangNetworkwideTrafficFlow2020,nie2023letc}. Consequently, changes in sensing topology can invalidate the learned model and necessitate costly retraining, underscoring limited resilience to multifaceted disturbances in the sensing environment.

Second, large-scale GTSI is inherently an underdetermined inverse problem with multiple sources of uncertainty. Unmeasured latent factors introduce intrinsic stochasticity ~\citep{R1-3}, while spatially sparse measurements permit multiple plausible solutions consistent with the observations. The interplay of these factors leads to complex uncertainty in inferred traffic states. Reliable quantification of this uncertainty is a prerequisite for trustworthy inference, as it enables decision-makers to assess the confidence of inferred states and act accordingly. However, existing work on uncertainty quantification (UQ) for traffic inference primarily targets forecasting at observed locations~\citep{cheng2024RecentAdvancesDeep}, whereas UQ for inferring states on uninstrumented segments remains limited. Moreover, we note that mobile sensing such as floating car  can traverse the entire road network, providing complementary traffic state observations, such as travel speed and floating vehicle counts, for segments unreachable by fixed detectors~\citep{tettamantiFleetDataBased2024,Badran2024JICV}. As these variables are correlated with traffic volume, they can offer additional observational constraints that help reduce the underdetermination. However, FCD generally has a low penetration rate, leaving its observations spatiotemporally sparse. To this end, fusing the two data sources offers a promising path toward jointly constraining GTSI; yet large-scale empirical evidence for such multi-source fusion remains scarce.

Third, GTSI can be naturally decomposed into three closely related subtasks: (1) forecasting at observed locations, (2) real-time state estimation on unobserved segments, and (3) forecasting on unobserved segments. A unified framework for these subtasks is attractive, as it can improve computational efficiency and avoid maintaining separate models. However, the three subtasks operate under different information regimes and therefore favor heterogeneous, and sometimes conflicting, spatiotemporal modeling assumptions. Specifically, forecasting at observed locations benefits from preserving location-specific variations and temporal dynamics, where estimation at unobserved locations relies more on spatial smoothing and shared spatial patterns. Forecasting at unobserved locations, lacking target-specific histories, relies more strongly on robust spatiotemporal common patterns.

However, the three subtasks impose partially conflicting requirements on the shared spatiotemporal representation. Spatially, forecasting at observed locations benefits from preserving location-specific variations, whereas estimation at unobserved locations relies more on suppressing such variations and exploiting smoother spatial patterns shared across locations. Temporally, forecasting into the future relies more on stable temporal trends and therefore favors suppressing short-term fluctuations, whereas real-time estimation can preserve more of the current temporal variability. Forecasting at unobserved locations combines both requirements, relying more heavily on spatially and temporally smooth patterns that generalize across locations and time. Prior work~\citep{zhou2025MoGERNNInductiveTraffic} suggests that these competing preferences can induce a seesaw effect across subtasks, where improving one subtask comes at the expense of another, making joint modeling non-trivial and still largely underexplored.

To address these challenges, we synergistically leverage floating car data (FCD) and sparse fixed detectors, formulate GTSI as a stochastic process, and propose the Task-Aware Attentive Neural Process (TA-ANP) for resilient and trustworthy inference of global traffic state. The key contributions of this work are fourfold.
\begin{itemize}

    \item We formulate each fixed-sensor configuration induced by multifaceted disturbances as a distinct inference context, and develop a context-conditioned meta-learning neural process that adapts to sensing-topology changes without expensive retraining. Experiments under compounded disturbances demonstrate resilience in disturbance absorption, performance recovery, and adaptation to unseen sensing configurations.
    
    \item We propose operational definitions of aleatoric and epistemic uncertainties in the context of GTSI and integrate neural processes with Monte Carlo (MC) Dropout to jointly capture both types of uncertainty. We further verify that the estimated uncertainty is well-calibrated with inference errors and effectively supports subsequent decision-making, such as fixed-sensor placement in multi-source data environments.

    \item We propose a cross-attention mechanism with task-specific query projections that enables adaptive aggregation of context information with differentiated spatiotemporal inductive biases for each subtask. This design effectively mitigates cross-task interference in the unified attention space, alleviating the seesaw effect across the three GTSI subtasks.
    
    \item We construct a metropolis-scale multi-source urban traffic dataset and benchmark TA-ANP against seven representative baselines. TA-ANP achieves best performance under both deterministic accuracy metrics and probabilistic scoring rules. For reproducibility, we have publicly released the processed datasets~\citep{zhou2026mmtd_fixed}, along with the implementation of TA-ANP and the data processing code on \href{https://github.com/qishenzhou/TA-ANP}{GitHub}.
\end{itemize}

The remainder of the paper is organized as follows: \cref{sec:LR} presents a review of the related literature. 
\cref{sec:PF} provides the formalization of the problem. \cref{sec:Method} details the proposed TA-ANP. \cref{sec:nm} reports on the performance and ablation results of the proposed method. \cref{sec:con} concludes the paper.

\section{Literature review} \label{sec:LR}

\subsection{Global traffic state inference (GTSI)}

The core challenge of GTSI lies in recovering network-wide traffic states from spatially sparse observations, which usually renders the problem undetermined. In practice, this calls for introducing additional constraints to narrow the feasible solution space. Depending on how such constraints are imposed, existing approaches can be broadly grouped into three categories: (1) imposing explicit mathematical structure into optimization modeling, (2) embedding spatiotemporal modeling assumptions or physical knowledge into the neural network design, and (3) incorporating side information to provide complementary observational constraints.

Tensor completion, as a representative of the first category, formulates state recovery as an optimization problem based on low-rank assumptions and further incorporates spatiotemporal regularization, such as graph Laplacian-based spatial smoothness and temporal autoregressive constraints, to narrow the solution space \citep{yu2016TemporalRegularizedMatrix, song2019TensorCompletionAlgorithms, zhangNetworkwideTrafficFlow2020,  Lei2022Kriging,nie2023letc}. However, these methods remain transductive, meaning any structural change in the sensing network requires re-optimization from scratch. Furthermore, incorporating additional data sources such as FCD typically necessitates non-trivial reformulation of the optimization problem structure.

Within the second strategy, Graph neural networks (GNNs) implicitly constrain through architectural inductive biases, introducing spatial similarity assumptions between neighboring nodes that allow observations at instrumented nodes to constrain the estimation at unobserved ones \citep{wu2021IGNNK, nie2023flowestimation, xu2025kits, zhou2025MoGERNNInductiveTraffic, zhou2025NetworkWidea}. Physics-informed machine learning (PIML) methods go further by explicitly embedding traffic flow models as learnable components into differentiable computational graphs \citep{zhan2017CitywideTrafficVolume, Yuan2022PIGP,Shi2021PIDL,wang2024Privacypreserving, guarda2024EstimatingNetworkFlow}. 
This explicit modeling comes with a strong dependence on detailed network specification (e.g., complex entry /exit structures )~\citep{ma2020Estimating}, and such methods may generalize poorly beyond the spatiotemporal range of the training data \citep{Kim2021DPM}. Compared to optimization-based approaches, the expressive power of deep learning better captures complex nonlinear spatiotemporal dependencies. However, under extremely sparse observations, even with architectural inductive biases and physical constraints the problem may be made technically well-posed, but the resulting estimates can remain low-fidelity and highly uncertain, motivating the incorporation of richer data sources.

Within the third strategy, a variety of auxiliary data have been leveraged, including FCD \citep{meng2017Citywide, tang2019JointModeling, zhangNetworkwideTrafficFlow2020, Xu2024multisource,ChenPeng2026}, social media check-in data \citep{shao2018LicensePlateRecognition}, turning rate \citep{xu2023AGNPNetworkwideShortterm}, and population information \citep{laraki2022LargeScale}, each offering a distinct dimension of constraint on the inference problem. Among these, FCD is particularly valuable as it directly captures traffic information and its availability has grown substantially with the proliferation of connected vehicles and mobile sensing. Nevertheless, FCD remains spatially sparse and temporally irregular, and the lack of publicly available large-scale datasets pairing FCD with fixed detector records has left large-scale empirical validation of multi-source fusion in GTSI nearly absent.

Orthogonal to the above constraint strategies, another gap in the literature is that the aforementioned studies predominantly focus on real-time state estimation on unobserved segments, whereas relatively few simultaneously address future state forecasting at both observed and unobserved locations~\citep{zhou2025MoGERNNInductiveTraffic, Roth2022Funs,Mei2023UIGNN}, and studies that jointly model all three GTSI subtasks remain rare.

\subsection{Resilience of cyber-physical traffic sensing systems}

In this work, we focus on the resilience of cyber-physical traffic sensing systems, i.e., the ability of the learning-based inference layer to remain dependable when the sensing configuration and data availability evolve due to disruptions in the physical sensing infrastructure. Drawing on a lifecycle view of resilience in learning-enabled systems~\citep{moskalenko2023Resilience}, we characterize resilience here in terms of disturbance absorption, recovery, and adaptation under evolving operating conditions.  From this perspective, a resilient model should (i) limit performance degradation under disruptions, (ii) recover as sensing capability is restored, and (iii) generalize to previously unseen sensor configurations without costly retraining. This framing motivates a closer look at modeling paradigms that explicitly support topology generalization and rapid adaptation.

Two methodological threads are particularly relevant to building resilient traffic sensing systems. Inductive GNNs, via sample-and-aggregate mechanisms, can generalize to unseen nodes and varying graph structures, providing a structural basis for accommodating topology changes~\citep{wu2021IGNNK, xu2025kits, zhou2025MoGERNNInductiveTraffic}. Meta-learning approaches, exemplified by neural processes, learn priors over inference functions from different contexts, enabling rapid adaptation to new sensing configurations without full retraining~\citep{garnelo2018NP, xu2023AGNPNetworkwideShortterm}.Although initial applications in GTSI, neither line has been systematically evaluated under compounded, multi-stage disturbances (e.g., damage, repair, and network expansion occurring in sequence), and their relative resilience has not been explicitly benchmarked under a unified experimental setting.

\subsection{Uncertainty quantification for traffic state inference}

Incorporating Uncertainty Quantification (UQ) into data-driven traffic state inference can support risk-aware decision-making in ITS \citep{cheng2024RecentAdvancesDeep}. In active control, UQ helps prioritize interventions for potential congestion with high confidence \citep{zechin2023Probabilistic}; in vehicle routing, it enables users to trade off ``fastest on average'' against ``most reliable travel time'' \citep{olivier2023Bayesian}. For the GTSI considered in this work, UQ provides an additional benefit: by producing uncertainty maps, it can guide active sensing~\citep{Mei2023UIGNN}, providing support for optimizing the placement of fixed-sensors and dynamically scheduling mobile sensing resources~\citep{Maljkovic2025, Xiong2025}.

In the deep learning literature, uncertainty is commonly decomposed into Aleatoric Uncertainty (AU) and Epistemic Uncertainty (EU). AU captures irreducible randomness arising from inherent stochasticity and measurement noise, whereas EU reflects reducible uncertainty attributable to limited data and model knowledge gaps; for comprehensive methodological taxonomies we refer readers to established surveys \citep{AbdarUQ2021, Gawlikowski2023SurveyUncertaintyDeep}.
In recent years, UQ has attracted increasing attention in data-driven TSI~\citep{cheng2024RecentAdvancesDeep}. Existing studies mainly incorporate UQ through several technical routes, including Monte Carlo dropout \citep{sengupta2024Bayesiana}, deep ensembles \citep{qian2023UncertaintyQuantificationTraffic}, and model designs that directly output prediction distributions \citep{li2022quantifying,xiong2026Unveiling}.  A closer look, however, reveals that most UQ efforts are concentrated on forecasting at observed locations, whereas UQ efforts in global inference settings involving state estimation and forecasting on unobserved segments remain underexplored. For example, some works employ inherently probabilistic mechanisms, such as generative adversarial networks \citep{xu2020GEGANNovelDeep} or KL-divergence-based objectives \citep{liu2024Networkwidea}, yet do not report or evaluate the resulting uncertainty estimates. To date, only a limited number of studies explicitly investigate UQ in global inference \citep{Lei2022Kriging,lei2024ConditionalDiffusionModel,xu2023AGNPNetworkwideShortterm,Mei2023UIGNN,wang2024KnowledgedataFusionOriented,ding2025Uncertainty}. To our knowledge, even fewer works explicitly model and quantify EU \citep{Mei2023UIGNN,ding2025Uncertainty}.This gap is critical for GTSI because it requires inferring states on many unobserved road segments, and the underlying state distributions can differ substantially across locations. If EU is not used to characterize what the model does not know in these unseen regions, the model can become overconfident and make erroneous inferences.

Moreover, recent surveys emphasize the need for deep probabilistic models in ITS that can update predictions and reassess uncertainty as new data arrive \citep{cheng2024RecentAdvancesDeep}. This requirement for dynamic UQ capability also resonates with the resilience-related challenges discussed in the last subsection.

Beyond producing an uncertainty estimate, verifying that it is trustworthy requires a separate body of evaluation methodology. A common practice is to report the prediction interval coverage probability (PICP), which evaluates coverage at a single nominal confidence level and is highly interpretable, yet a trivially wide, uninformative interval can satisfy nominal coverage without conveying useful information~ \citep{gneitingStrictlyProperScoring2007}. The quantile interval coverage error (QICE) extends this check across the full range of quantile intervals rather than a single confidence level, offering a more comprehensive diagnosis of calibration~\citep{Han2022QICE}. However, calibration, whether assessed at a single level or across the full quantile range, must nonetheless be complemented by sharpness, i.e., how concentrated the predictive distribution is, and strictly proper scoring rules such as the continuous ranked probability score (CRPS) provide a principled way to jointly evaluate both~ \citep{gneitingStrictlyProperScoring2007}. Existing studies on traffic state inference typically adopt only a single metric for evaluation, and few works systematically report all three metrics jointly. 

\section{Problem formulation} \label{sec:PF}

We study Global Traffic State Inference (GTSI) in a road network with a road-segment set $\mathcal{V}$ over a historical window $\mathcal{H}_h=\{1,\dots,T\}$ and a future window $\mathcal{H}_f=\{T+1,\dots,T+H\}$. We denote the full spatiotemporal domain as $\mathcal{S}=\mathcal{V}\times(\mathcal{H}_h\cup\mathcal{H}_f)$. Each spatiotemporal point $(\nu,t)$ is represented by an identity vector $\mathbf{x}(\nu,t)$ that concatenates (i) temporal features $\mathbf{\tau}_t$ (e.g., time of day and day of week), (ii) static road attributes $\mathbf{s}_\nu$ (e.g., road class and number of lanes), and (iii) dynamic road representations $\mathbf{e}_{\nu,T}$ derived from FCD (e.g., volume and median speed of floating cars). Here, $\mathbf{e}_{\nu,T}$ is obtained by encoding the FCD observations of road segment $\nu$ over the historical window $\mathcal{H}_h$. It is shared across all time points within the current inference horizon and is updated as the historical window advances. The traffic state at $(\nu,t)$ is denoted by $y(\nu,t)\in\mathbb{R}^+$ and is instantiated as traffic volume in this study. Observations are available only on $\mathcal{S}_\mathcal{C}=\mathcal{V}_o\times\mathcal{H}_h$, where $\mathcal{V}_o \subsetneq \mathcal{V}$ denotes road segments equipped with fixed sensors, forming the context set $\mathcal{C}=\{(\mathbf{x}(\nu,t),y(\nu,t)) \mid (\nu,t) \in \mathcal{S}_\mathcal{C}\}$. The target set $\mathcal{T}$ includes all remaining points in $\mathcal{S}$, covering (1) spatial imputation at time $t\le T$ (estimation at unobserved locations), (2) forecasting at observed locations and (3) forecasting at unobserved locations. Our primary objective is to approximate the conditional distribution $p(Y_{\mathcal{T}}\mid X_{\mathcal{T}},\mathcal{C})$.

To achieve this, we view traffic states as realizations from a stochastic process evaluated at $\mathbf{x}(\nu,t)$:
\begin{equation}
    y(\nu,t)=f(\mathbf{x}(\nu,t))+\epsilon,\ \epsilon\sim\mathcal{N}(0,\sigma^2),
    \label{eq:sp}
\end{equation}
where $f$ is a random function (i.e., a sample path) drawn from the process prior and $\epsilon$ represents the combined effect of sensor measurement noise and unmodeled intrinsic systemic randomness. Although we adopt a Gaussian assumption ($\epsilon \sim \mathcal{N}(0, \sigma^2)$) for mathematical tractability, this noise distribution can be refined to domain-specific forms based on the characteristics of the real-world network and observation in future extensions.

\begin{figure}
    \centering
    \includegraphics[width=0.9\linewidth]{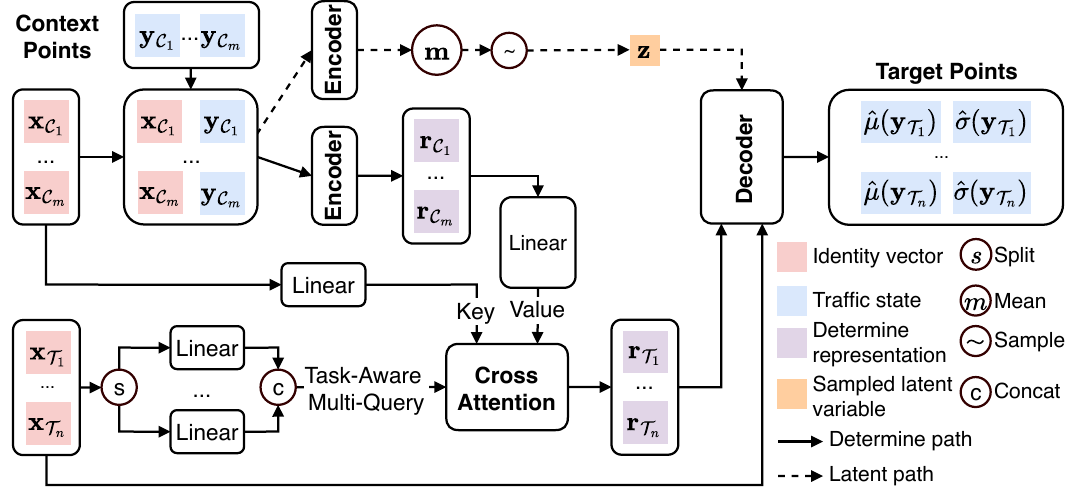}
    \caption{The overview of Task-Aware Attentive Neural Process (TA-ANP).}
    \label{fig:TA-ANP}
\end{figure}

\section{Methodology} \label{sec:Method}

A natural Bayesian instantiation of the above stochastic-process view is the Gaussian Process (GP). The fundamental premise of a GP lies in encoding prior knowledge about correlations between different locations via a covariance function (i.e., kernel function) $k(\mathbf{x}, \mathbf{x}')$. Leveraging these correlations, the GP performs an optimal weighted combination of observed information from the context set to infer the state of any target point. This characteristic allows GP to flexibly fit complex dynamic traffic patterns. Furthermore, the analytic posterior distribution provided by its Bayesian framework inherently supports uncertainty quantification, aligning perfectly with the objectives of this work. However, these benefits come at a prohibitive computational cost: exact GP inference scales as $\mathcal{O}((n+m)^3)$ in the total number of inputs, where $n$ and $m$ denote the sizes of the context (observed) set and the target (query) set, respectively. This cubic scaling makes global inference on large-scale urban road networks impractical. 

To address this, we adopt Neural Processes (NPs) \citep{garnelo2018NP} as our foundational framework. Through amortized inference with neural networks, NPs reduce the per-episode computational complexity to $\mathcal{O}(n+m)$. To better capture the multi-task nature of global traffic inference, we further propose Task-Aware Attentive Neural Processes (TA-ANP), built on the NP architecture. An overview of TA-ANP is shown in Figure \ref{fig:TA-ANP}. We first introduce the NP backbone and then describe how attention and a task-aware multi-query design improve performance.

\subsection{Stochastic process modeling via Neural Processes}

The Neural Process (NP) family is commonly instantiated as Conditional Neural Processes (CNP) \citep{garnelo2018CNP} and Latent Neural Processes (LNP) \citep{garnelo2018NP}. Given a context set $\mathcal{C}=\{(\mathbf{x}_i,y_i)\}_{i=1}^{n}$ and target inputs $\mathbf{x}_\mathcal{T}$, both variants share three building blocks:

\begin{itemize}
    \item \textbf{Encoder}: maps each context pair $(\mathbf{x}_i,y_i)\in\mathcal{C}$ to a representation $\mathbf{r}_i$, i.e., $\mathbf{r}_i=h_{\theta}(\mathbf{x}_i,y_i)$.
    \item \textbf{Aggregator}: combines $\{\mathbf{r}_i\}_{i=1}^{n}$ into a single global summary $\mathbf{R}$ using a permutation-invariant operator (e.g., mean pooling), i.e., $\mathbf{R}=\mathrm{AGG}(\{\mathbf{r}_i\})$.
    \item \textbf{Decoder}: conditions on $\mathbf{x}_\mathcal{T}$ and the aggregated representation to parameterize the predictive distribution at targets, i.e., $p(y_\mathcal{T} \mid \mathbf{x}_\mathcal{T}, \mathbf{R})$.
\end{itemize}

The key difference lies in how the prior function is represented. The CNP is purely deterministic and decodes directly from the aggregated representation $\mathbf{R}$. In contrast, LNP introduces a stochastic latent path by using $\mathbf{R}$ to parameterize $p(\mathbf{z}\mid\mathcal{C})$ and marginalizing over $\mathbf{z}$ at prediction time, i.e., $p(y_\mathcal{T}\mid\mathbf{x}_\mathcal{T},\mathcal{C})=\int p(y_\mathcal{T}\mid\mathbf{x}_\mathcal{T},\mathbf{z})\,p(\mathbf{z}\mid\mathcal{C})\,d\mathbf{z}$. This stochasticity improves uncertainty modeling and enables diverse function samples.

Despite this difference, both CNP and LNP typically compress the entire context set into a fixed-dimensional vector, creating an information bottleneck that obscures the heterogeneous contributions of individual context points. For traffic fields with strong local spatial autocorrelation and for GTSI scenarios that couple multiple prediction subtasks, such indiscriminate compression can wash out critical local signals and blur task-specific cues, degrading predictive performance.

\subsection{Task-aware attentive neural process}

To alleviate the global-summary bottleneck of CNP/LNP, this study adopts Attentive Neural Processes (ANP) \citep{kim2019ANP} as the backbone architecture, which augments the LNP-style latent path with an attention-based deterministic path implemented by multi-head cross-attention. In this mechanism, the target point identifier vector serves as the query ($\mathbf{Q}$), while the context point identifier vectors and traffic states serve as keys ($\mathbf{K}$) and values ($\mathbf{V}$), respectively. By computing attention weights, the mechanism adaptively aggregates context information $\mathbf{r}_\mathcal{C}$ to generate a target-specific representation $\mathbf{r}_\mathcal{T}$. The decoder simultaneously receives the latent variable $\mathbf{z}$ from the latent path and $\mathbf{r}_\mathcal{T}$ from the deterministic path, thereby enhancing the representation of local autocorrelation.

As previously discussed, global inference essentially encompasses three subtasks. From a signal processing perspective, these three tasks exhibit fundamentally different dependencies on spatiotemporal frequency domain features:

\begin{itemize}
    \item \textbf{Real-time estimation at unobserved locations}: Lacking historical priors for specific nodes, the model must interpolate values for unobserved locations based on spatial homogeneity. The inference accuracy relies on extracting spatial low-frequency components (i.e., global trends and smooth evolution). In this context, excessive spatial high-frequency details not only fail to aid inference but can lead to overfitting. Therefore, the mechanism design must prioritize spatial smoothing and the capture of common features.

    \item \textbf{Forecasting at observed locations}: This task possesses complete historical observation sequences for the target nodes. To achieve high-fidelity forecasting, the model must retain spatial high-frequency components to accurately characterize the local fluctuations unique to the node, distinguishing it from its neighborhood. Temporally, the model relies on temporal low-frequency components to capture robust evolutionary trends.

    \item \textbf{Forecasting for unobserved locations}: As the most challenging task, it lacks historical references and requires the capture of robust temporal trends. This demands that the model focus on extracting dual spatiotemporal low-frequency components. The inference process must filter out high-frequency interference to achieve robust generalized deduction.
\end{itemize}

Standard ANP employs a single shared linear projection $\mathbf{W}^Q$ to generate query vectors for all tasks (i.e., $\mathbf{Q}'=\mathbf{Q}\mathbf{W}^Q$). This parameter-sharing mechanism forces the model to trade off conflicting generalization requirements (e.g., spatial low-frequency vs. spatial high-frequency) within a unified feature space, making it prone to feature interference and negative transfer between tasks.

To address these challenges, this study proposes a Task-Aware Multi-Query Mechanism (TAMQM). While maintaining shared projections for keys ($\mathbf{K}$) and values ($\mathbf{V}$), this mechanism defines independent linear projection matrices for the three subtasks:
\begin{equation}
\mathbf{Q}'_{\mathrm{task}} = \mathbf{Q}_{\mathrm{task}} \mathbf{W}^Q_{\mathrm{task}}, \quad \mathrm{where } \mathbf{W}^Q_{\mathrm{task}} \in \{ \mathbf{W}^Q_{\mathrm{s}}, \mathbf{W}^Q_{\mathrm{t}}, \mathbf{W}^Q_{\mathrm{st}} \}
\end{equation}
Through this design, TAMQM allows each subtask to learn a distinct query representation and attention pattern, thereby accommodating their heterogeneous spatiotemporal dependencies and reducing interference caused by a fully shared attention space.

\subsection{Uncertainty modeling in global traffic inference}

Uncertainty is commonly decomposed into Epistemic Uncertainty (EU) and Aleatoric Uncertainty (AU). EU reflects uncertainty due to limited knowledge and is, in principle, reducible, whereas AU accounts for variability that is typically treated as irreducible (e.g., inherent randomness and observation noise) \citep{hullermeier2021Aleatoric}. Distinguishing and modeling these two components is important for achieving well-calibrated predictions and supporting safety-critical downstream decisions.

At the same time, the philosophical boundary between EU and AU is not absolute. Under idealized assumptions of complete knowledge and unlimited resources, one may argue that even uncertainty often regarded as aleatoric could be reduced and thus subsumed into EU \citep{vrouwenvelder2003Keynote,faber2005Treatment}. Such a view, however, is of limited practical value: it would imply arbitrarily complex models and make the EU/AU distinction dependent on the decision-maker's information state. Therefore, following the pragmatic framework of \citet{kiureghian2009Aleatory}, we adopt the following operational definitions of EU and AU tailored to the GTSI setting:
\begin{itemize}
    \item \textbf{AU} refers to the uncertainty induced by observation noise and unmodeled factors that perturb system dynamics and are difficult or intentionally not eliminated from the outset.

    \item \textbf{EU} refers to uncertainty caused by limited knowledge, including the non-uniqueness of plausible solutions under sparse sensing and potential model misspecification under data distribution shifts. In principle, EU can be reduced by acquiring additional observations.
\end{itemize}
These definitions provide an operational basis for allocating sensing resources and guiding model improvement. Next, we describe how TA-ANP models these two uncertainty components.

TA-ANP models AU through its probabilistic decoder, which outputs the parameters of a Gaussian predictive distribution, i.e., $p(y_\mathcal{T} \mid \mathbf{r}_\mathcal{T}, \mathbf{z}) = \mathcal{N}(\mu_\mathcal{T}, \sigma_\mathcal{T}^2)$. The predicted variance $\sigma_\mathcal{T}^2$ directly captures the aleatoric component attributable to observation noise and unmodeled factors \citep{kendall2017What}.

TA-ANP captures EU by sampling multiple prediction functions from the global latent posterior $\mathbf{z} \sim q(\mathbf{z}\mid\mathcal{C})$ that are consistent with the available context observations. The resulting dispersion in function space reflects solution non-uniqueness under sparse sensing, thereby characterizing EU due to sensor sparsity. However, because TA-ANP uses fixed model parameters, it cannot represent EU arising from model misspecification under distribution shifts. As a result, in out-of-distribution scenarios, the model may fail to communicate its knowledge gaps and produce overconfident predictions.

To improve EU quantification, we further cast TA-ANP in a Bayesian neural network framework. As exact Bayesian inference is intractable, we follow \citet{gal2016Dropout} and employ Monte Carlo Dropout (MC Dropout) to obtain an approximate variational posterior over model parameters. Concretely, we insert dropout layers into the encoder and decoder and perform $K$ stochastic forward passes at inference time, generating a set of predictive samples $\{(\mu_i,\sigma_i^2)\}_{i=1}^{K}$. Using the law of total variance \citep{kendall2017What,valdenegro2022}, the predictive variance at a target point is decomposed as
\begin{equation}
\mathrm{Var}(y^*) \approx \underbrace{\frac{1}{K} \sum_{i=1}^K \sigma_i^2}_{\mathrm{AU}} + \underbrace{\frac{1}{K} \sum_{i=1}^K (\mu_i - \bar{\mu})^2}_{\mathrm{EU}},
\label{eq:agg}
\end{equation}
where $\bar{\mu} = \frac{1}{K} \sum_{i=1}^K \mu_i$ is the predictive mean of the ensemble. The first term estimates AU by averaging predictive variances, whereas the second term estimates EU via the dispersion of predictive means.

\subsection{Model training}

The model is trained by maximizing the log-likelihood of the target observations $\mathbf{y}_{\mathcal{T}}$ conditioned on the context set $\mathcal{C}$. Because the model contains continuous latent variables $\mathbf{z}$, the marginal likelihood involves an intractable integral. We therefore maximize the evidence lower bound (ELBO), and equivalently minimize the negative-ELBO objective
\begin{equation}
    \mathcal{L}(\theta,\phi)=\mathcal{L}_{\mathrm{NLL}}+\beta\,\mathcal{L}_{\mathrm{KL}},
\end{equation}
where $\theta$ and $\phi$ denote the parameters of the encoder and decoder, respectively, and $\beta$ controls the strength of the latent-space regularization.

The first term, $\mathcal{L}_{\mathrm{NLL}}$, corresponds to the predictive negative log-likelihood. Assuming an independent Gaussian predictive distribution $p(y_i\mid\mathbf{x}_i,\mathbf{z},\mathcal{C})=\mathcal{N}(\mu_i,\sigma_i^2)$, we have
\begin{equation}
    \mathcal{L}_{\mathrm{NLL}}
    =-\mathbb{E}_{q(\mathbf{z}\mid\mathcal{C})}\!\left[\sum_{(\mathbf{x}_i,y_i)\in\mathcal{T}}\log\mathcal{N}(y_i\mid\mu_i,\sigma_i^2)\right]
    \approx \sum_{(\mathbf{x}_i,y_i)\in\mathcal{T}}\left(\frac{(y_i-\mu_i)^2}{2\sigma_i^2}+\log\sigma_i\right)+\mathrm{const.}
\end{equation}
This term encourages accurate fitting while penalizing overconfident predictions through the variance-dependent weighting.

The second term, $\mathcal{L}_{\mathrm{KL}}$, is a Kullback--Leibler (KL) regularizer that aligns the variational posterior inferred from the context with that inferred from a larger target set. Following the NP training procedure summarized by \citet{le2018empirical}, we randomly sample a batch-specific target set $\mathcal{T}'$ from the observations and then subsample a context set $\mathcal{C}'\subset\mathcal{T}'$. The KL term is defined as
\begin{equation}
    \mathcal{L}_{\mathrm{KL}}=D_{\mathrm{KL}}\bigl(q(\mathbf{z}\mid\mathcal{C}')\,\|\,q(\mathbf{z}\mid\mathcal{T}')\bigr).
\end{equation}

\begin{table}[!t]
    \centering
    \caption{Summary statistics of the urban-road subset used in this paper.}
    \label{tab:mmtd_stats}
    \begin{tabular}{lp{0.68\linewidth}}
        \hline
        \textbf{Item} & \textbf{Value}\\
        \hline
        Study area (Madrid urban) & Above 200 km$^2$ \\
        Road segments ($N$) & 26{,}166\\
        Fixed-sensor-equipped road segments & 2{,}371\\
        Target variable & Traffic volume measured by fixed sensors \\
        Fixed-sensor data missing rate & 9.76\% \\
        TomTom penetration rate & 2\%-10\% \\
        Time resolution & 15 min\\
        Study period & Aug.1st 2024 - Aug.30th 2024\\
        Attributes & Fixed-sensor flow, FCD flow, FCD speed, static road attributes (OSM-derived), betweenness and closeness centrality (segment level) \\
        \hline
    \end{tabular}
\end{table}

\begin{figure}[]
\centering
\begin{subfigure}[t]{0.48\textwidth}
  \centering
  \includegraphics[width=\linewidth]{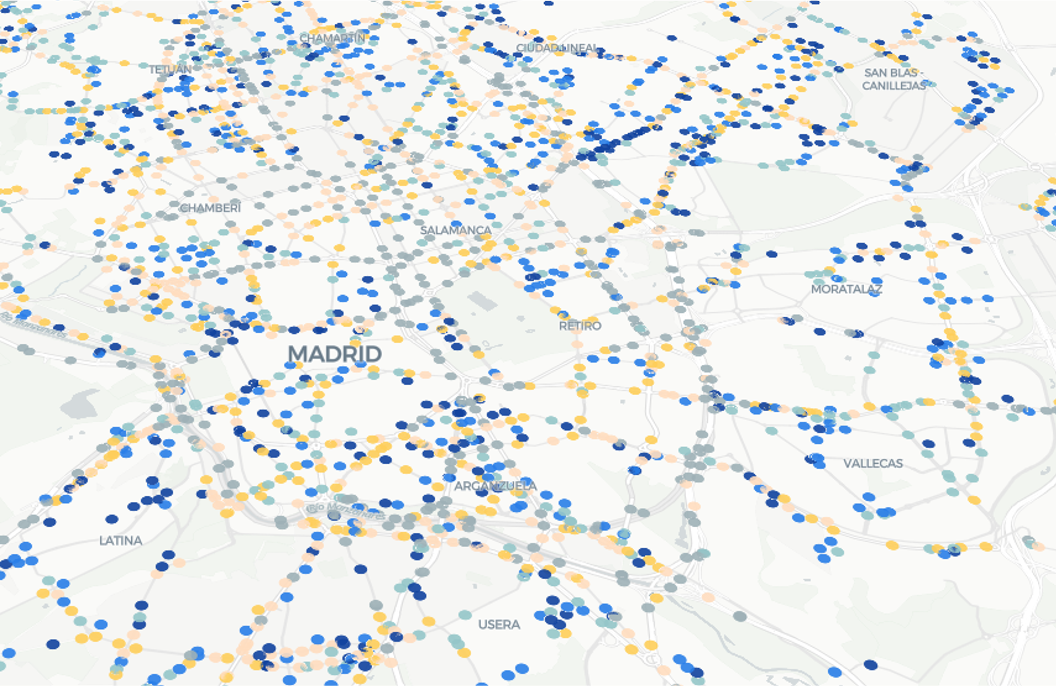}
  \caption{Spatial distribution of fixed sensors in MMTD.}
  \label{fig:mmtd_sensors}
\end{subfigure}\hfill
\begin{subfigure}[t]{0.49\textwidth}
  \centering
  \includegraphics[width=\linewidth]{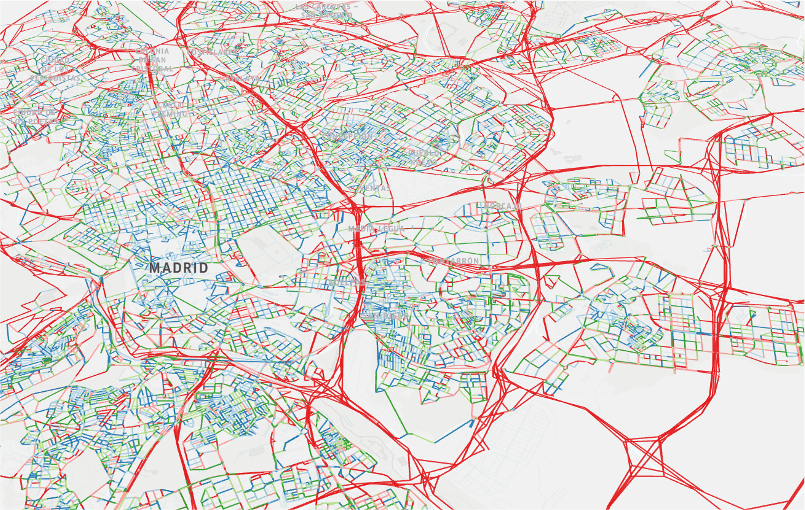}
  \caption{TomTom data coverage in MMTD.}
  \label{fig:mmtd_tomtom}
\end{subfigure}
\caption{Map view of MMTD}
\end{figure}

\section{Numerical experiments} \label{sec:nm}

This section evaluates TA-ANP for metropolis-scale GTSI on an urban road network at the road-segment level. To support this evaluation, we constructed the Metropolitan Multi-Source Traffic Dataset (MMTD) by integrating (i)fixed-loop sensor measurements released by the Madrid City Council \citep{MadridFLow}, (ii) floating-car traffic statistics provided by TomTom \citep{TomTom}, (iii) an OpenStreetMap (OSM) road network \citep{osmnx}, (iv) and lane-count annotations released by the Madrid City Council \citep{MadridLane}. MMTD is produced through a preprocessing pipeline that aligns heterogeneous data layers with physical road segments, normalizes segment granularity, consolidates co-located sensors, and cleans raw measurements. The resulting observations are aggregated at a 15 min resolution. TomTom traffic information is updated at approximately a half-minute scale, which is substantially finer than the 15 min temporal resolution adopted in this study; therefore,  we do not explicitly model the end-to-end transmission and processing latency of the FCD feed. We focus on traffic volume inference over the 2,371 fixed-sensor-equipped road segments within the 26,166-segment urban network, for which ground-truth traffic flow is available. TomTom statistics and OSM-derived road attributes, such as road class and segment length, as well as graph-based network measures, serve as auxiliary observations. Dataset details are reported in \cref{tab:mmtd_stats}, and the spatial coverage of fixed sensors and FCD is illustrated in \cref{fig:mmtd_sensors} and \cref{fig:mmtd_tomtom}, respectively. For reproducibility, we release the processed fixed-sensor data together with the road-network files and node features \citep{zhou2026mmtd_fixed}. We also provide code for integrating TomTom data and loading multi-source observations, while TomTom data cannot be redistributed due to licensing constraints \citep{ZhouMMTDCode}.

A set of experiments is conducted to answer the following questions.
\begin{itemize}
    \item \textbf{Q1: Overall performance.} (\cref{sec:q1})
    Across the three GTSI subtasks, how does TA-ANP compare with representative baselines in terms of point accuracy and UQ quality? To what extent do the estimated uncertainties align with the inference errors?

    \item \textbf{Q2: Uncertainty-guided sensor placement.} (\cref{sec:q2})
    Can the state inference uncertainty quantified by TA-ANP effectively guide fixed sensor placement in multi-source sensing (fixed sensors and probe-vehicle data), thereby reducing greater state inference errors with fewer sensor inputs?

    \item \textbf{Q3: Resilience under multifaceted sensing disturbances.} (\cref{sec:q3})
    How resilient is TA-ANP to sensing disturbances induced by network damage, repair, and incremental deployment?

    \item \textbf{Q4: Impact of fixed sensor deployment density.} (\cref{sec:q4})
    As the availability of fixed sensors decreases, how do the predictive accuracy and UQ performance of TA-ANP's evolve and how does its degradation curve differ from those of the baselines?

    \item \textbf{Q5: Gains from FCD and penetration effects.} (\cref{sec:q5})
    How do different information components in FCD contribute to GTSI performance? How do probe-vehicle penetration rate and inherent spatiotemporal heterogeneity influence TA-ANP's inference performance?

    \item \textbf{Q6: Ablation of the core components of TA-ANP.} (\cref{sec:q6})
    To what extent do TAMQM and MC Dropout contribute to the overall performance and UQ quality of TA-ANP, and are they essential components of the model?
\end{itemize}

\subsection{Experimental setup}
\paragraph{Baseline methods.} In benchmarking our approach, we selected seven representative models categorized into spatiotemporal GNNs—AGCRN \citep{LeiAGCRN20}, STCAGCN \citep{nie2023flowestimation}, DGAE \citep{zhou2025NetworkWidea}, and MoGERNN \citep{zhou2025MoGERNNInductiveTraffic}—and uncertainty-aware models—UIGNN \citep{Mei2023UIGNN}, DeepSTUQ \citep{qian2023UncertaintyQuantificationTraffic}, and DistPred \citep{Liang2025}. To ensure a fair comparison while preserving the core design principles of each baseline method, we have made the necessary extensions to their feature-processing modules, enabling them to leverage the multi-source information provided by the MMTD dataset.


\paragraph{Evaluation metrics.}To comprehensively assess model performance, we establish evaluation metrics from two dimensions: deterministic prediction and probabilistic prediction. For all metrics, the original missing values in the dataset are excluded from the calculations. To evaluate the accuracy of deterministic predictions, we employ five widely used metrics, including Mean Absolute Error (MAE), Root Mean Square Error (RMSE), Symmetric Mean Absolute Percentage Error (SMAPE), Relative Root Mean Square Error (RRMSE) and the Coefficient of Determination ($R^2$). Furthermore, probabilistic calibration quality is assessed via Continuous Ranked Probability Score (CRPS) \citep{matheson1976CRPS}, Prediction Interval Coverage Probability (PICP)~\citep{R0}, Mean Prediction Interval Width (MPIW)~\citep{R0}, and Quantile Interval Calibration Error (QICE) \citep{Han2022QICE}. Standard deterministic metrics (MAE, RMSE, and $R^2$) follow their conventional definitions and are omitted for brevity; the definitions of SMAPE, RRMSE, CRPS, PICP, MPIW and QICE are provided in Appendix \ref{sec:metrics-define}.

\paragraph{Implementation details.} The proposed TA-ANP model is implemented using PyTorch 2.6.0 and is deployed and trained on an NVIDIA RTX 4090 GPU. Architectural hyperparameter settings for the model components are summarized in Appendix \ref{sec:A2}. For uncertainty estimation, MC Dropout is performed with 30 stochastic forward passes in the main experiments. Each forward pass samples dropout masks independently and draws a new latent sample $\mathbf{z}$ from the learned latent distribution. The resulting predictive distributions are aggregated following the procedure described in \cref{eq:agg} to obtain the final predictive mean and total predictive variance, which together define the final predictive distribution used for probabilistic evaluation. Both the input history length and the prediction length are set to 4 time steps (i.e., 1 hour). The training process utilizes the AdamW optimizer for parameter updates. Throughout the temporal dimension, the entire dataset is partitioned into training and test sets in a 6:4 ratio. To avoid overfitting, a portion of the data is randomly sampled from the time periods corresponding to the training set to form a validation set, which is used in conjunction with an early stopping strategy. During the sample construction phase, a sliding window with a step of 1 is applied to extract the time series. Unless otherwise specified, experiments uniformly sampled 40\% of the 2,371 road segments equipped with a fixed sensor (949 in total) as observed locations, while the remaining 60\% (1,422 in total) are treated as unobserved locations by excluding their fixed-sensor measurements from the model inputs. The measurements at these unobserved locations are used solely as ground truth for evaluation. For the principal comparison in \cref{tab:over_all}, this spatial partitioning procedure is independently repeated five times, and the reported results are the mean and sample standard deviation across the five partitions. The spatial distributions of the observed and unobserved locations are presented in Appendix~\ref{app:sensor_partitions}.

\begin{table}[!tb]
\centering
\caption{Comparison of Traffic Flow Inference Performance Across Different Models over Five Independently Generated Sensor Partitions. Values are reported as mean with the sample standard deviation shown in smaller type. Bold and underlined text indicate the best and second-best mean results, respectively. The \textit{Improvement} column quantifies the gain of TA-ANP over the best baseline, computed from unrounded mean values. Metrics marked with $\downarrow$ indicate that lower values correspond to better model performance, while those marked with $\uparrow$ indicate the opposite. The PICP metric, marked with an asterisk (*), is considered better the closer it is to the ideal coverage level of 95\%. MPIW is interpreted jointly with PICP; it is therefore not ranked and has no percentage improvement.}
\label{tab:over_all}
\begin{tabularx}{\textwidth}{llXXXXXXXX|X}
\toprule
\multicolumn{2}{c}{\small \textbf{Model}} & \small \textbf{AGCRN} & \small \textbf{STCA\newline GCN} & \small \textbf{DGAE} & \small \textbf{MoGE\newline RNN} & \small \textbf{UIGNN} & \small \textbf{Deep\newline STUQ} & \small \textbf{DistPred} & \small \textbf{TA-ANP} & \small \textbf{Improve} \\
\midrule
\multirow{9}{*}{\rotatebox{90}{\shortstack{Estimation for \\ Unobserved Location}}} & MAE $\downarrow$ & 31.44{\tiny$\pm$0.43} & 34.05{\tiny$\pm$0.34} & 33.28{\tiny$\pm$1.05} & 31.05{\tiny$\pm$0.45} & 31.71{\tiny$\pm$0.57} & \uline{30.83{\tiny$\pm$0.99}} & 35.68{\tiny$\pm$0.71} & \textbf{25.96{\tiny$\pm$0.53}} & \textit{+15.77} \\
 & RMSE $\downarrow$ & 45.33{\tiny$\pm$0.56} & 47.57{\tiny$\pm$0.40} & 46.91{\tiny$\pm$1.45} & \uline{42.84{\tiny$\pm$1.14}} & 44.79{\tiny$\pm$1.09} & 43.09{\tiny$\pm$1.32} & 50.30{\tiny$\pm$1.19} & \textbf{37.79{\tiny$\pm$0.96}} & \textit{+11.77} \\
 & SMAPE $\downarrow$ & \uline{36.66{\tiny$\pm$0.66}} & 39.13{\tiny$\pm$0.86} & 38.26{\tiny$\pm$1.58} & 36.81{\tiny$\pm$0.72} & 37.35{\tiny$\pm$1.09} & 36.99{\tiny$\pm$1.37} & 40.77{\tiny$\pm$1.07} & \textbf{29.95{\tiny$\pm$0.59}} & \textit{+18.30} \\
 & RRMSE $\downarrow$ & 46.35{\tiny$\pm$0.31} & 48.64{\tiny$\pm$0.53} & 47.98{\tiny$\pm$1.59} & \uline{43.80{\tiny$\pm$1.07}} & 45.81{\tiny$\pm$1.38} & 44.07{\tiny$\pm$1.58} & 51.43{\tiny$\pm$1.32} & \textbf{38.65{\tiny$\pm$1.15}} & \textit{+11.77} \\
 & R$^2$ $\uparrow$ & 0.55{\tiny$\pm$0.011} & 0.50{\tiny$\pm$0.011} & 0.51{\tiny$\pm$0.027} & \uline{0.60{\tiny$\pm$0.021}} & 0.56{\tiny$\pm$0.023} & 0.59{\tiny$\pm$0.025} & 0.44{\tiny$\pm$0.028} & \textbf{0.69{\tiny$\pm$0.019}} & \textit{+15.05} \\
 & CRPS $\downarrow$ & / & / & / & / & 23.62{\tiny$\pm$0.65} & \uline{22.65{\tiny$\pm$0.80}} & 76.84{\tiny$\pm$4.07} & \textbf{18.83{\tiny$\pm$0.42}} & \textit{+16.87} \\
 & PICP$^*$ & / & / & / & / & \uline{0.87{\tiny$\pm$0.025}} & 0.86{\tiny$\pm$0.025} & 0.24{\tiny$\pm$0.033} & \textbf{0.93{\tiny$\pm$0.013}} & \textit{/} \\
 & MPIW & / & / & / & / & 136.81{\tiny$\pm$8.25} & 122.41{\tiny$\pm$5.93} & 95.93{\tiny$\pm$3.60} & 120.60{\tiny$\pm$8.17} & \textit{/} \\
 & QICE $\downarrow$ & / & / & / & / & \uline{0.011{\tiny$\pm$0.0020}} & 0.016{\tiny$\pm$0.0049} & 0.076{\tiny$\pm$0.0020} & \textbf{0.007{\tiny$\pm$0.0003}} & \textit{+39.31} \\
\midrule
\multirow{9}{*}{\rotatebox{90}{\shortstack{Forecasting at \\ Unobserved Location}}} & MAE $\downarrow$ & 33.71{\tiny$\pm$0.88} & 35.78{\tiny$\pm$1.14} & 34.54{\tiny$\pm$0.85} & \uline{31.11{\tiny$\pm$1.05}} & 34.69{\tiny$\pm$0.35} & 34.02{\tiny$\pm$1.00} & 37.98{\tiny$\pm$0.96} & \textbf{26.83{\tiny$\pm$0.58}} & \textit{+13.77} \\
 & RMSE $\downarrow$ & 48.22{\tiny$\pm$1.15} & 50.24{\tiny$\pm$1.43} & 48.19{\tiny$\pm$0.95} & \uline{43.05{\tiny$\pm$1.95}} & 48.23{\tiny$\pm$0.81} & 47.18{\tiny$\pm$1.40} & 52.79{\tiny$\pm$1.51} & \textbf{38.76{\tiny$\pm$1.05}} & \textit{+9.96} \\
 & SMAPE $\downarrow$ & 39.35{\tiny$\pm$1.13} & 40.79{\tiny$\pm$1.39} & 39.62{\tiny$\pm$1.26} & \uline{36.65{\tiny$\pm$0.90}} & 41.37{\tiny$\pm$0.50} & 41.38{\tiny$\pm$1.91} & 43.32{\tiny$\pm$1.01} & \textbf{30.97{\tiny$\pm$0.65}} & \textit{+15.49} \\
 & RRMSE $\downarrow$ & 49.39{\tiny$\pm$0.90} & 51.45{\tiny$\pm$1.34} & 49.36{\tiny$\pm$1.06} & \uline{44.09{\tiny$\pm$1.85}} & 49.40{\tiny$\pm$1.10} & 48.33{\tiny$\pm$1.65} & 54.07{\tiny$\pm$1.62} & \textbf{39.71{\tiny$\pm$1.25}} & \textit{+9.95} \\
 & R$^2$ $\uparrow$ & 0.49{\tiny$\pm$0.025} & 0.44{\tiny$\pm$0.034} & 0.49{\tiny$\pm$0.019} & \uline{0.59{\tiny$\pm$0.038}} & 0.49{\tiny$\pm$0.019} & 0.51{\tiny$\pm$0.027} & 0.39{\tiny$\pm$0.035} & \textbf{0.67{\tiny$\pm$0.020}} & \textit{+13.14} \\
 & CRPS $\downarrow$ & / & / & / & / & 25.72{\tiny$\pm$0.43} & \uline{25.41{\tiny$\pm$1.15}} & 75.20{\tiny$\pm$3.06} & \textbf{19.44{\tiny$\pm$0.48}} & \textit{+23.47} \\
 & PICP$^*$ & / & / & / & / & \uline{0.91{\tiny$\pm$0.013}} & 0.81{\tiny$\pm$0.063} & 0.30{\tiny$\pm$0.033} & \textbf{0.92{\tiny$\pm$0.015}} & \textit{/} \\
 & MPIW & / & / & / & / & 162.26{\tiny$\pm$7.59} & 116.64{\tiny$\pm$16.50} & 107.29{\tiny$\pm$4.61} & 122.52{\tiny$\pm$8.31} & \textit{/} \\
 & QICE $\downarrow$ & / & / & / & / & \uline{0.010{\tiny$\pm$0.0017}} & 0.023{\tiny$\pm$0.0078} & 0.073{\tiny$\pm$0.0026} & \textbf{0.007{\tiny$\pm$0.0003}} & \textit{+33.14} \\
\midrule
\multirow{9}{*}{\rotatebox{90}{\shortstack{Forecasting at \\ Observed Location}}} & MAE $\downarrow$ & 18.34{\tiny$\pm$0.78} & 19.17{\tiny$\pm$0.40} & 19.42{\tiny$\pm$0.93} & \uline{17.39{\tiny$\pm$0.64}} & 18.55{\tiny$\pm$0.14} & 19.29{\tiny$\pm$0.91} & 17.46{\tiny$\pm$0.15} & \textbf{15.55{\tiny$\pm$0.19}} & \textit{+10.58} \\
 & RMSE $\downarrow$ & 26.19{\tiny$\pm$1.19} & 27.40{\tiny$\pm$0.45} & 26.66{\tiny$\pm$1.30} & 24.50{\tiny$\pm$0.90} & 26.49{\tiny$\pm$0.24} & 27.19{\tiny$\pm$1.27} & \uline{24.45{\tiny$\pm$0.22}} & \textbf{22.16{\tiny$\pm$0.35}} & \textit{+9.34} \\
 & SMAPE $\downarrow$ & 22.58{\tiny$\pm$0.76} & 26.25{\tiny$\pm$1.18} & 25.32{\tiny$\pm$1.32} & 22.64{\tiny$\pm$1.01} & 23.66{\tiny$\pm$0.32} & 25.46{\tiny$\pm$1.56} & \uline{22.47{\tiny$\pm$0.06}} & \textbf{19.81{\tiny$\pm$0.18}} & \textit{+11.86} \\
 & RRMSE $\downarrow$ & 27.05{\tiny$\pm$1.40} & 28.29{\tiny$\pm$0.37} & 27.53{\tiny$\pm$1.39} & 25.30{\tiny$\pm$0.96} & 27.35{\tiny$\pm$0.43} & 28.08{\tiny$\pm$1.32} & \uline{25.24{\tiny$\pm$0.29}} & \textbf{22.88{\tiny$\pm$0.37}} & \textit{+9.35} \\
 & R$^2$ $\uparrow$ & 0.85{\tiny$\pm$0.015} & 0.83{\tiny$\pm$0.006} & 0.84{\tiny$\pm$0.016} & 0.87{\tiny$\pm$0.011} & 0.84{\tiny$\pm$0.005} & 0.84{\tiny$\pm$0.015} & \uline{0.87{\tiny$\pm$0.003}} & \textbf{0.89{\tiny$\pm$0.004}} & \textit{+2.60} \\
 & CRPS $\downarrow$ & / & / & / & / & \uline{13.46{\tiny$\pm$0.12}} & 13.96{\tiny$\pm$0.74} & 71.33{\tiny$\pm$1.62} & \textbf{11.17{\tiny$\pm$0.13}} & \textit{+17.00} \\
 & PICP$^*$ & / & / & / & / & \uline{0.94{\tiny$\pm$0.007}} & 0.89{\tiny$\pm$0.037} & 0.10{\tiny$\pm$0.023} & \textbf{0.96{\tiny$\pm$0.006}} & \textit{/} \\
 & MPIW & / & / & / & / & 93.90{\tiny$\pm$3.89} & 80.91{\tiny$\pm$8.43} & 81.54{\tiny$\pm$3.42} & 82.54{\tiny$\pm$3.19} & \textit{/} \\
 & QICE $\downarrow$ & / & / & / & / & \uline{0.009{\tiny$\pm$0.0007}} & 0.015{\tiny$\pm$0.0065} & 0.073{\tiny$\pm$0.0354} & \textbf{0.008{\tiny$\pm$0.0011}} & \textit{+11.00} \\
\bottomrule
\end{tabularx}
\end{table}

\subsection{Overall performance evaluation for global traffic flow inference (Q1)} \label{sec:q1}

As summarized in \cref{tab:over_all}, TA-ANP achieves the best mean performance across the three GTSI subtasks under both deterministic and probabilistic metrics over five independently generated sensor partitions. The relatively small standard deviations across partitions indicate that the performance of all methods is stable with respect to sensor selection, while the advantage of TA-ANP is preserved under the repeated-partition evaluation. Among deterministic baselines, MoGERNN is generally the strongest competitor, whereas DeepSTUQ and UIGNN provide the strongest probabilistic baselines depending on the metric. Notably, although DistPred performs well for point-wise traffic forecasting \citep{Liang2025}, its PICP in global inference involving unobserved locations remains far below the nominal level of 0.95, indicating poorly calibrated uncertainty estimates.

For estimation at unobserved locations, TA-ANP consistently achieves the best mean performance across all deterministic metrics. Relative to the strongest baseline for each metric, it reduces prediction errors by 11.77\%--18.30\% and improves $R^2$ by 15.05\%. For uncertainty quantification, TA-ANP reduces CRPS and QICE by 16.87\% and 39.31\%, respectively, while achieving a PICP of 0.93 with an MPIW of 120.60. Notably, UIGNN produces a substantially wider interval (MPIW$=136.81$) but still attains poorer coverage (PICP$=0.87$). This suggests that its coverage error is not primarily caused by insufficient interval width, but may instead reflect bias or other forms of distributional miscalibration; its higher QICE further supports this interpretation. In contrast, TA-ANP achieves coverage closer to the nominal level with a more concentrated predictive interval.

For the more challenging task of forecasting at unobserved locations, MoGERNN remains the strongest deterministic baseline. TA-ANP further reduces prediction errors by 9.95\%--15.49\% relative to the strongest baseline for each metric and increases $R^2$ from 0.59 to 0.67. At the probabilistic level, TA-ANP reduces CRPS and QICE by 23.47\% and 33.14\%, respectively, while achieving a PICP of 0.92 with an MPIW of 122.52. UIGNN, despite producing a much wider interval (MPIW$=162.26$), achieves slightly poorer coverage (PICP$=0.91$) and a higher QICE. This again indicates that simply widening the prediction interval does not resolve distributional miscalibration. TA-ANP therefore provides a more favorable balance between calibration and sharpness.

For forecasting at observed locations, all models perform comparatively well. TA-ANP nevertheless remains the best-performing model across the deterministic metrics, reducing prediction errors by 9.34\%--11.86\% relative to the strongest baseline for each metric and increasing $R^2$ to 0.89. For probabilistic forecasting, TA-ANP reduces CRPS by 17.00\% and QICE by approximately 11\%, while achieving a PICP of 0.96 with an MPIW of 82.54. Compared with UIGNN, TA-ANP achieves similarly near-nominal coverage with a substantially narrower interval; compared with DeepSTUQ, a slightly wider interval is accompanied by markedly improved coverage and calibration.

Overall, the repeated-partition results confirm that TA-ANP consistently provides accurate and well-calibrated inference across heterogeneous GTSI settings, with particularly pronounced gains for estimation and forecasting at unobserved locations. The joint PICP--MPIW comparison further shows that improved coverage is not simply achieved by widening prediction intervals: several baselines produce wider intervals yet remain farther from the nominal coverage level, indicating broader distributional miscalibration, as also reflected by their higher QICE values. The small variation across independently generated sensor partitions further indicates that these improvements are not dependent on a particular sensor configuration.

\begin{figure}[]
    \centering
    \includegraphics[width=0.9\linewidth]{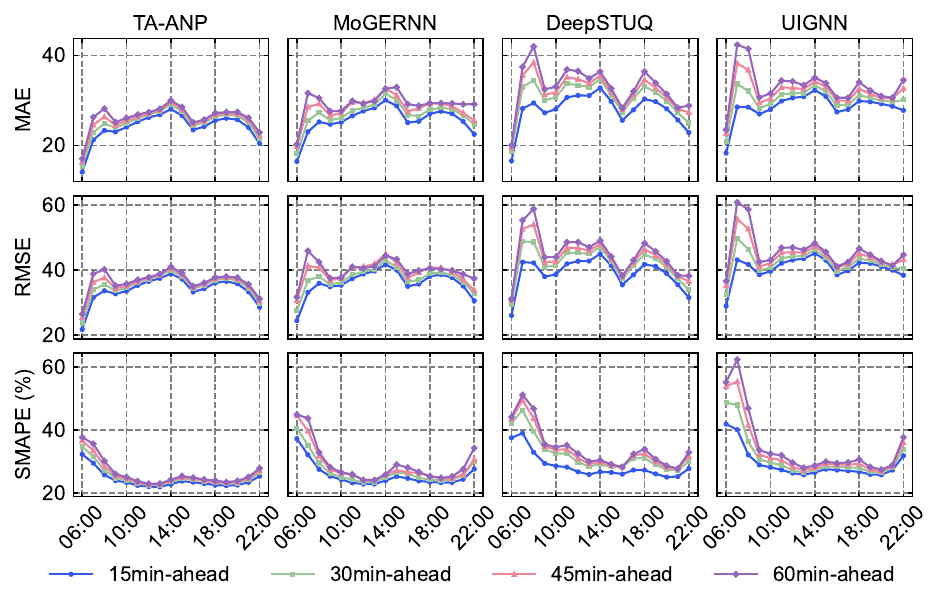}
    \caption{Diurnal evolution of network-wide prediction errors (averaged over all 2{,}371 locations). Columns correspond to TA-ANP and three baselines (MoGERNN, DeepSTUQ, and UIGNN); rows report MAE, RMSE, and SMAPE. Curves show forecast horizons from 15 to 60 minutes.}
    \label{fig:error_time}
\end{figure}
\begin{figure}[]
    \centering
    \includegraphics[width=0.9\linewidth]{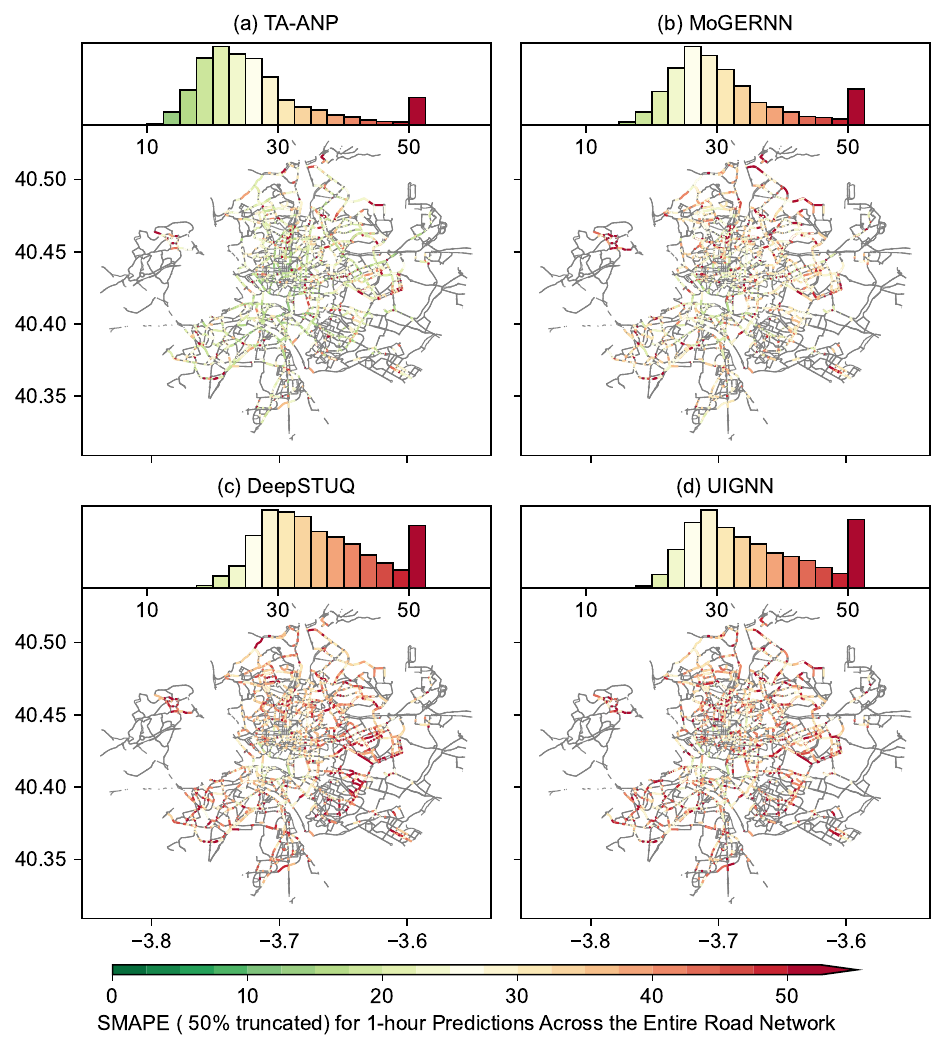}
    \caption{Spatial distribution and histogram of SMAPE(\%) for 1-hour-ahead prediction over the entire road network. Each subfigure reports the SMAPE histogram (top; values above 50\% are truncated) and the corresponding spatial heatmap (bottom; green to red indicates low to high error).}
    \label{fig:map_smape}
\end{figure}

\subsubsection{Deterministic prediction performance of TA-ANP}

To assess performance under complex diurnal traffic dynamics, \cref{fig:error_time} summarizes network-wide error profiles across the day for forecast horizons from 15 to 60 minutes. As the horizon increases, the errors of DeepSTUQ and UIGNN grow rapidly, whereas TA-ANP and MoGERNN remain comparatively stable. Notably, TA-ANP at the 60-minute horizon achieves MAE and RMSE that are lower than those of several baselines even at the 30-minute horizon, highlighting its robustness for longer-term prediction.

\cref{fig:map_smape} further compares models in terms of their spatial error patterns for 1-hour-ahead prediction. TA-ANP exhibits a markedly more favorable error distribution, with most SMAPE values below 30\%, compared with below 40\% for MoGERNN and substantial mass in the 40\%--50\% range for DeepSTUQ and UIGNN. Spatially, TA-ANP mitigates clusters of high-error regions in the urban core, which is critical given the greater complexity of traffic and the demand for control in downtown areas. Errors are generally higher near the network boundary, likely due to sparse fixed sensors and limited probe-vehicle coverage, which reduces the available information for inference in these regions.

\begin{figure}[]
    \centering
    \includegraphics[width=0.9\linewidth]{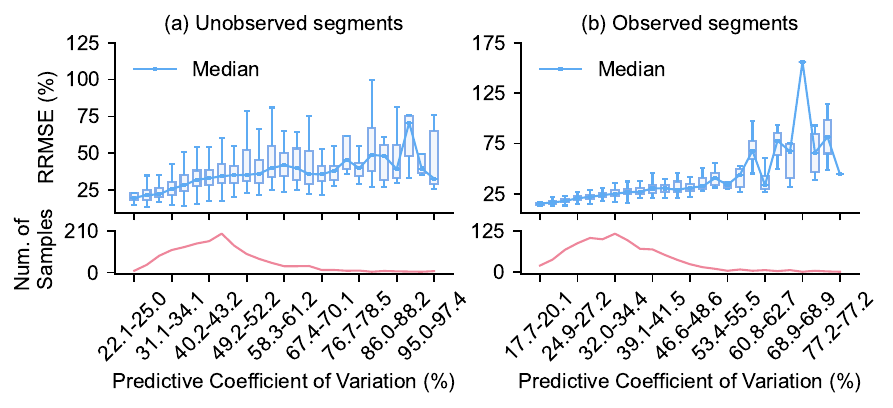}
    \caption{Relationship between the predictive coefficient of variation (PCV, \%) and relative root mean square error (RRMSE, \%) for TA-ANP (1-hour-ahead global inference). Panels correspond to (a) unobserved and (b) observed road segments. In each panel, the upper plot shows the distribution of RRMSE within each PCV bin (box plots) with the median connected by a line, and the lower plot reports the corresponding sample counts.}
    \label{fig:cv_vs_rrmse}
\end{figure}
\begin{figure}[]
    \centering
    \includegraphics[width=0.9\linewidth]{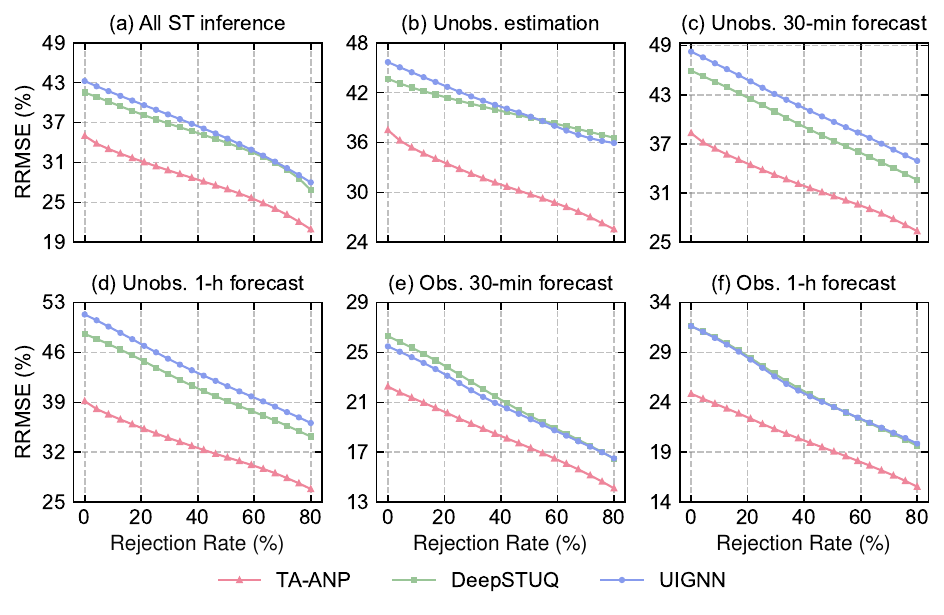}
    \caption{Error--rejection curves showing RRMSE(\%) versus rejection rate(\%) when discarding samples with the highest uncertainty. (a) All spatiotemporal (ST) points inference. (b) Unobserved (Unobs.) estimation. (c) Unobs. 30-min forecast. (d) Unobs. 1-h forecast. (e) Observed (Obs.) 30-min forecast. (f) Obs. 1-h forecast.}
    \label{fig:screen}
\end{figure}

\subsubsection{Predictive performance of TA-ANP in terms of predictive distributions}
This section evaluates whether uncertainty quantified by TA-ANP is well calibrated to its prediction errors. We consider global inference results with a prediction of 1 hour in advance for all locations and examine the relationship between the predictive coefficient of variation (PCV) and RRMSE. PCV is defined as

\begin{equation}
PCV = \frac{\hat{\sigma}}{\hat{\mu}} \times 100\%
\end{equation}
where $\hat{\sigma}$ and $\hat{\mu}$ denote the standard deviation and mean of the predictive distribution, respectively.

\cref{fig:cv_vs_rrmse} relates the predictive coefficient of variation (PCV) to RRMSE for both unobserved and observed road segments. Overall, the results suggest strong consistency between uncertainty and error. For unobserved segments, roughly 80\% of samples lie in the PCV range of 22\%--55\%, and the median RRMSE exhibits a stable positive association with PCV. For observed segments, which benefit from richer historical observations, the PCV--RRMSE relationship is approximately linear for PCV $<50\%$. Although the association becomes weak for very high PCV ($>60\%$), this regime contains only a negligible fraction of samples and therefore has little impact on aggregate performance. From an operational perspective, PCV can serve as a practical proxy for prediction reliability. For example, signal control can adjust the conservativeness of control parameters based on predicted uncertainty, and routing algorithms can prioritize links with higher certainty to improve robustness and efficiency.

To further assess whether uncertainty can be used to screen out high-risk cases, \cref{fig:screen} reports error--rejection curves based on the \emph{predictive coefficient of variation} (PCV). TA-ANP consistently yields the lowest RRMSE across all scenarios, with a smoothly decreasing error curve as the rejection rate increases, indicating that its uncertainty estimates effectively prioritize high-error samples. In the real-time estimation task for unobserved segments (subfigure~(b)), the performance gap between TA-ANP and the baselines widens as more uncertain samples are removed, underscoring TA-ANP's advantage in probabilistic inference for unobserved nodes. Overall, TA-ANP delivers not only accurate point predictions, but also highly informative uncertainty estimates for risk-aware screening.

\begin{figure}[]
    \centering
    \includegraphics[width=0.9\linewidth]{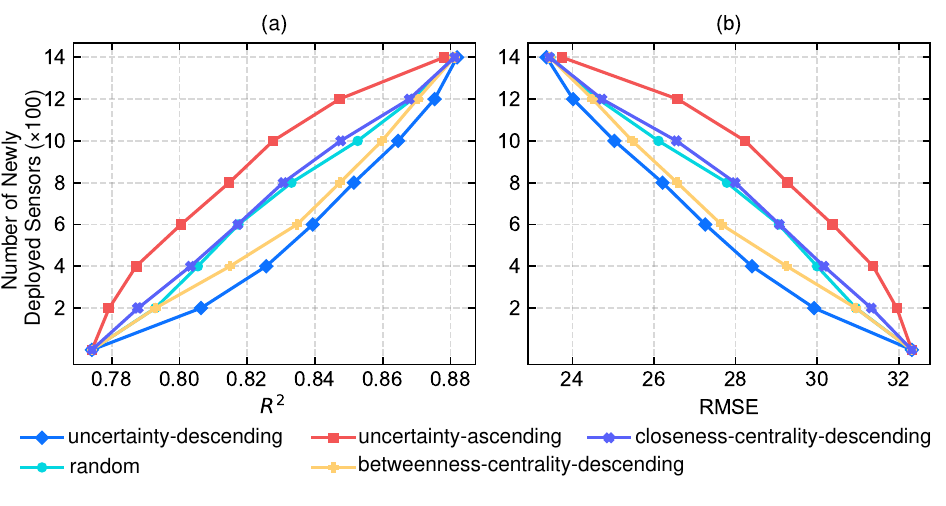}
    \caption{Fixed-sensor placement efficiency under different proxy metrics in an incremental deployment setting. The y-axis indicates the number of newly deployed sensors ($\times 100$). Panels report prediction performance versus deployment cost in terms of (a) $R^2$ and (b) RMSE. Curves correspond to uncertainty-guided deployment (descending/ascending), graph-centrality heuristics (betweenness/closeness), and random placement.}
    \label{fig:active_sensing}
\end{figure}

\subsection{Active sensor placement guided by estimated uncertainty (Q2)} \label{sec:q2}

Optimizing sensor placement under budget constraints has long been a core issue in ITS. Most prior work assumes a single-source setting dominated by fixed detectors, and typically (i) enforces network observability via algebraic flow-conservation constraints or (ii) relies on coverage-driven heuristics (e.g., maximum coverage) to rank candidate locations \citep{owais2022Traffic}. With the advent of FCD, however, the logic of fixed-sensor deployment shifts: fixed detectors should be sited to optimally complement mobile observations whose coverage is extensive yet sparse and spatiotemporal heterogeneous. Accordingly, this section investigates whether the overall uncertainty quantified by TA-ANP can serve as an informative criterion for fixed-sensor site selection in a multi-source fusion setting.

We consider an incremental deployment scenario with 60\% initially unobserved road segments. The number of fixed sensors increases from 948 to 2348, reducing the unobserved proportion to 1\% at the end. We compare five placement strategies: (1) \emph{uncertainty-descending} (ours), which prioritizes segments with higher TA-ANP uncertainty; (2) \emph{uncertainty-ascending}, used as a negative control; (3) \emph{betweenness-centrality-descending}; (4) \emph{closeness-centrality-descending}; and (5) \emph{random} placement as an uninformed baseline.

\cref{fig:active_sensing} shows that TA-ANP's uncertainty provides clear guidance for sensor placement: uncertainty-descending consistently achieves the best performance per deployed sensor, whereas uncertainty-ascending performs worst, confirming a strong association between this uncertainty metric and inference error. The advantage is most pronounced in the early, resource-limited stage: to reach $R^2=0.825$, uncertainty-descending requires only about 400 sensors, saving approximately 20\% (\,100 sensors\,) relative to the best-performing baseline (betweenness centrality) and about 43\% (\,300 sensors\,) relative to random placement. These results indicate that, in multi-source fusion scenarios, uncertainty-guided deployment can allocate limited sensing resources more efficiently and accelerate improvements in network-level inference accuracy.

\begin{figure}
    \centering
    \includegraphics[width=0.9\linewidth]{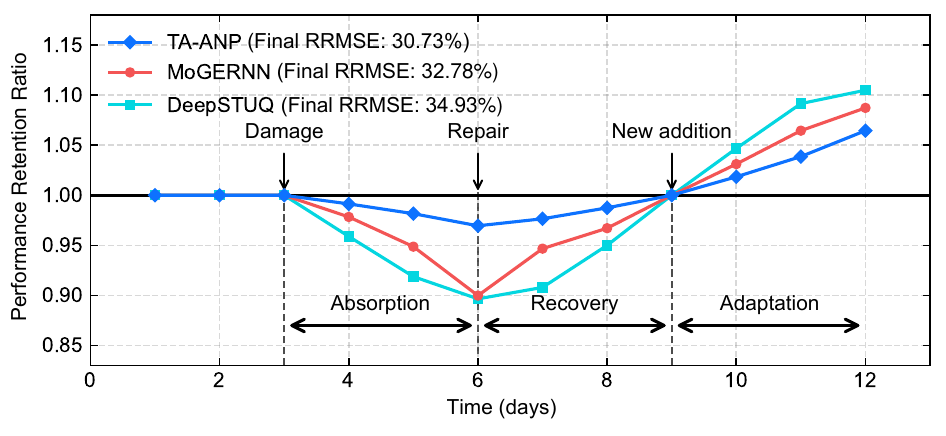}
    \caption{Resilience of flow inference under a Damage--Repair--Addition lifecycle of the fixed-sensor network. The y-axis reports the performance retention ratio, defined as $(1/\mathrm{RRMSE})/(1/\mathrm{RRMSE}_{\mathrm{base}})$; thus $y=1$ indicates the disturbance-free baseline. Vertical dashed lines separate the three stages. Legends report final RRMSE values at day~12.}
    \label{fig:risilient}
\end{figure}

\subsection{Model resilience under multifaceted sensing disturbances (Q3)}\label{sec:q3}

In this case study, we assess the resilience of TA-ANP to multifaceted sensing disturbances by simulating a \emph{Damage--Repair---Addition} lifecycle in the fixed-sensor network. Starting from a 60\% initially unobserved ratio, we perturb the network over three stages: (i) \emph{damage} (days 3--6), where 100 sensors are removed per day; (ii) \emph{repair} (days 6--9), where 100 previously damaged sensors are restored per day following a first-damaged-first-repaired (FIFO) rule; and (iii) \emph{addition} (days 9--12), where 100 new sensors are deployed per day. Damaged and newly added sensors are selected uniformly at random.

To allow fair comparison across stages with different effective network sizes, we report a performance retention ratio in Fig.~\ref{fig:risilient}. The results show that TA-ANP outperforms the baselines along three key dimensions of resilience:

\begin{itemize}
\item \textbf{Disturbance absorption}: TA-ANP exhibits the strongest disturbance absorption during the damage stage (days 3--6). At the most severe damage point, DeepSTUQ and MoGERNN drop to about 0.90 in retention ratio, whereas TA-ANP remains at 0.97, indicating that it can better exploit the remaining spatial dependencies to mitigate performance loss.
\item \textbf{Recovery}: During the repair stage (days 6--9), TA-ANP recovers steadily toward the disturbance-free baseline, suggesting that it can promptly incorporate restored data streams and re-establish disrupted spatiotemporal dependencies.
\item \textbf{Adaptability to network expansion}: In the addition stage (days 9--12), all methods surpass the benchmark ($y>1.0$) and enter a positive-gain regime. Although TA-ANP shows a smaller relative slope due to its stronger baseline, it still achieves the lowest final error, reaching an RRMSE of 30.73\% on day 12.
\end{itemize}

Overall, TA-ANP maintains high accuracy under sensor loss, recovers rapidly as sensors are repaired, and continues to benefit from incremental deployments, demonstrating robust adaptability to previously unseen sensing configurations.

\begin{figure}[]
    \centering
    \includegraphics[width=0.9\textwidth]{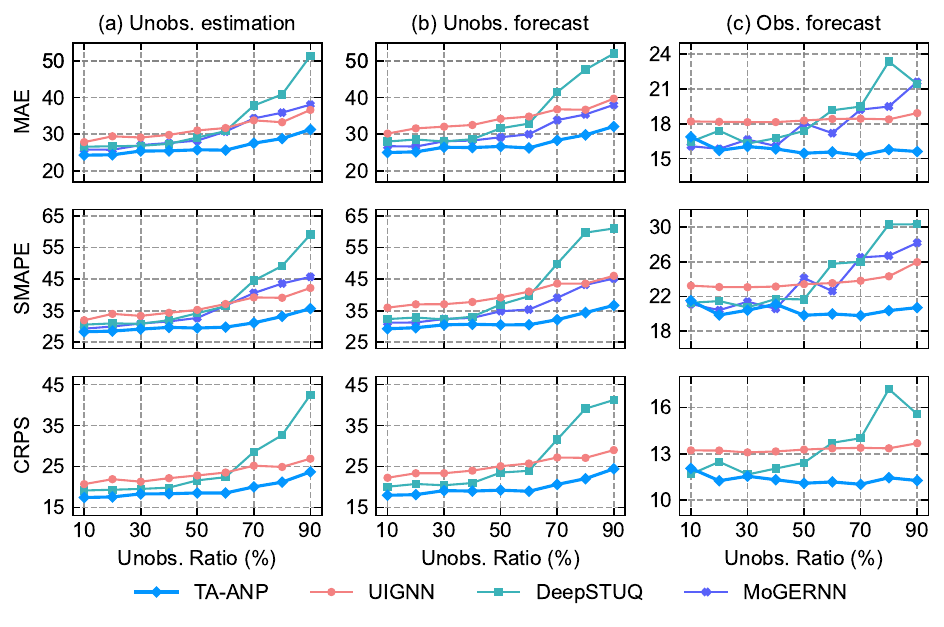}
    \caption{Comparison of model performance under different unobserved ratios (Unobs. Ratio, \%). Columns correspond to (a) Unobs. estimation, (b) Unobs. forecasting, and (c) Obs. forecasting, where forecasting results are averaged over multiple prediction horizons. Rows report MAE, SMAPE, and CRPS.}
    \label{fig:error_vs_density}
\end{figure}

\subsection{Impact of fixed sensor deployment density (Q4)} \label{sec:q4}
To examine how fixed-sensor deployment density affects inference quality, we simulate a spectrum of sensing regimes by varying the proportion of unobserved locations from 10\% to 90\% and conduct systematic comparisons. For clarity, we report results for three representative baselines (DeepSTUQ, MoGERNN, and UIGNN), which show strong overall performance in \cref{tab:over_all}. \cref{fig:error_vs_density} summarizes the results across tasks. Overall, TA-ANP yields lower errors than the baselines in nearly all settings; below we focus on the more practically relevant tasks of estimation and forecasting at unobserved locations.

For estimation and forecasting at unobserved locations (\cref{fig:error_vs_density}a--b), we observe a clear transition around a 50\%--60\% unobserved ratio: beyond this range, most methods degrade more rapidly, with DeepSTUQ exhibiting the steepest deterioration. Before this transition, MoGERNN and DeepSTUQ are the strongest baselines among deterministic and probabilistic models, respectively; at higher missing ratios, UIGNN becomes competitive and overtakes DeepSTUQ. Quantitatively, in estimation at unobserved locations, increasing the unobserved ratio from 10\% to 90\% raises TA-ANP's MAE by 7.00, whereas the best baseline (MoGERNN) increases by 12.27 (about $1.75\times$ larger), indicating stronger robustness to sparse sensing. This robustness translates into improved deployment cost-effectiveness. For example, in multi-step forecasting at unobserved locations (averaged over all prediction horizons), with MAE=30 as an acceptable threshold, TA-ANP maintains performance up to 80\% unobserved, while MoGERNN exceeds the threshold at 60\%, implying that comparable accuracy can be achieved with roughly 20\% fewer fixed sensors (474 sensors in our setup). A similar pattern holds for probabilistic inference: with CRPS=20 as the target, TA-ANP remains below the threshold up to 70\% unobserved, whereas DeepSTUQ reaches CRPS=21.60 already at 50\% unobserved, again suggesting an approximate 20\% reduction in required deployments. Taken together, these results highlight TA-ANP's superior robustness and cost-effectiveness for both deterministic prediction and UQ under sparse fixed sensing.

\begin{figure}[]
    \centering
    \includegraphics[width=0.9\textwidth]{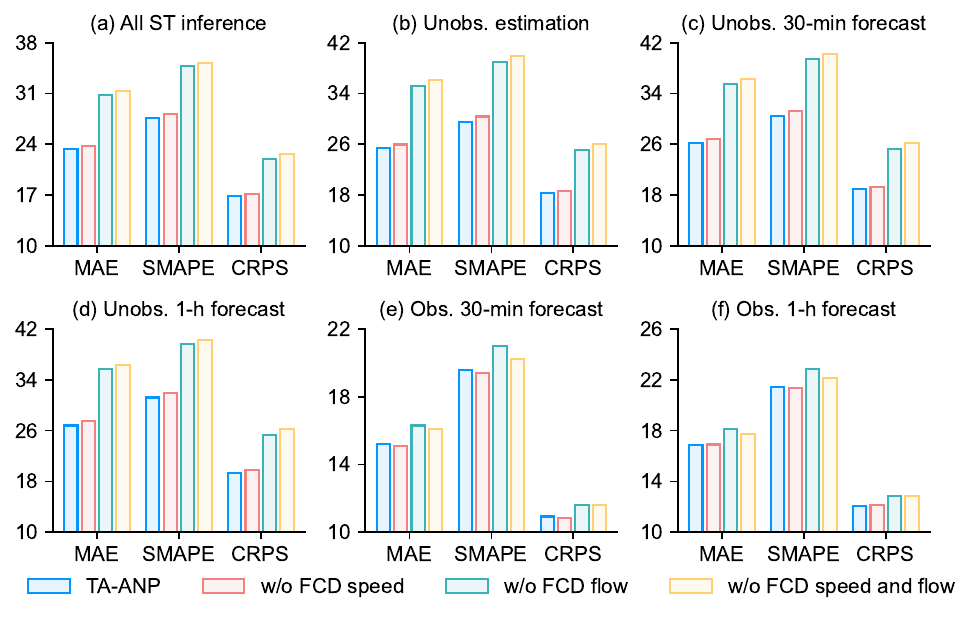}
    \caption{Ablation on the contribution of floating-car data (FCD) in network-wide flow inference. Subplots (a)--(f) follow the same six scenarios as Fig.~\ref{fig:screen}. Each subplot reports MAE, SMAPE, and CRPS for TA-ANP and variants removing FCD speed and/or flow.}
    \label{fig:tom_benifit}
\end{figure}

\begin{figure}[]
    \centering
    \includegraphics[width=0.9\linewidth]{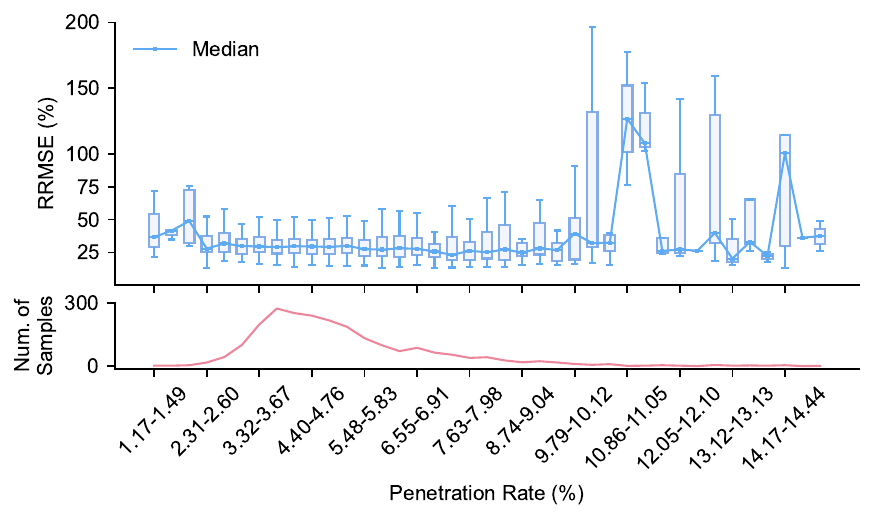}
    \caption{Variation in network-wide flow forecasting error (RRMSE, \%) across floating-car data (FCD) penetration-rate bins (1-hour horizon). The upper panel shows the distribution of RRMSE within each bin (box plots) with the median connected by a line, and the lower panel reports the corresponding sample counts.}
    \label{fig:prvs_rrmse}
\end{figure}

\subsection{Gains from FCD and penetration effects (Q5)} \label{sec:q5}

To quantify the benefit of FCD for network-wide traffic flow inference, we conducted controlled experiments with 60\% of road segments treated as unobserved. As shown in \cref{fig:tom_benifit}, we remove FCD speed, FCD flow, or both and compare the resulting changes across tasks. Two main observations emerge.

First, FCD flow contributes substantially more than speed. For real-time estimation on unobserved segments (\cref{fig:tom_benifit}b), removing speed only slightly increases MAE (25.38 $\rightarrow$ 25.95). By contrast, removing flow increases MAE to 35.16, a relative performance degradation of 38.5\%. This trend is consistent across all tasks, indicating that FCD flow is the dominant driver of performance gains.

Second, the reliance on FCD differs markedly between observed and unobserved segments. For segments with historical fixed-sensor observations (subfigures~e and~f), removing FCD leads to a modest increase in MAE (\,<15\%). In contrast, for inference on unobserved segments, the MAE can increase by up to 35.7\% (subfigure~d). This highlights the indispensable role of FCD in covering the blind spots of sparse fixed sensors. Additional comparisons further show that, with FCD, using only 10\% fixed-sensor coverage can outperform the setting without FCD even when fixed-sensor coverage reaches 90\%, underscoring the importance of FCD for inferring states on unobserved segments.

In our dataset, FCD mainly comes from commercial vehicles, resulting in low penetration and pronounced spatial heterogeneity, which can challenge GTSI. We therefore evaluate robustness under different penetration levels by stratifying segments into penetration-rate bins (\cref{fig:prvs_rrmse}). Note that the penetration rate is a segment-wise time average induced by data aggregation, rather than a controlled experimental variable.

As shown in \cref{fig:prvs_rrmse}, across the penetration range 2.3\%--9.5\% (about 92\% of segment samples), the model remains stable: the median RRMSE stays within 20\%--26\%, and the mean within 30\%--35\%. These results suggest that the proposed method is effective under low penetration and is robust to spatiotemporal variations in penetration. The larger volatility at higher penetration levels is mainly attributable to the small sample size in those bins (sample imbalance), rather than a reduced benefit of higher penetration for network-wide state inference.

\begin{table}[]
\renewcommand\arraystretch{1.5}
\caption{Results of the ablation experiment on key modules of TA-ANP.Values in parentheses indicate the percentage change in performance relative to the full TA-ANP model, where negative values indicate degraded performance.}\label{tab:ablation}
\begin{tabularx}{\textwidth}{llYYYYYY}
\toprule
\multicolumn{2}{c}{\textbf{Model}}                                                                  &\textbf{MAE}$\downarrow$      & \textbf{SMAPE}$\downarrow$ & \textbf{R²}$\uparrow$ & \textbf{CRPS}$\downarrow$      & \textbf{PICP}$^*$    & \textbf{QICE}$\downarrow$    \\ \midrule
\multirow{4}{*}[-3.5ex]{\rotatebox{90}{\shortstack{Estimation for \\ Unobserved Location}}} & TA-ANP  & \textbf{25.63} & \textbf{29.80} & \textbf{0.70 }& 1\textbf{8.51} & \textbf{0.94} &\textbf{ 0.007}       \\ \cmidrule{2-8}
                                                                          & w/o TAMQM & \perfdown{27.92}{8.93} & \perfdown{32.26}{8.26} & \perfdown{0.63}{10.00} & \perfdown{20.82}{12.48} & 0.84 & \perfdown{0.012}{71.43}        \\
                                                                          & w/o MC Dropout & \perfdown{29.73}{16.00} & \perfdown{34.71}{16.48} & \perfdown{0.59}{15.71} & \perfdown{22.71}{22.69} & 0.78 & \perfdown{0.019}{171.43}        \\
                                                                          & MC Dropout $\rightarrow$ Dropout & \perfdown{27.32}{6.59} & \perfdown{31.41}{5.40} & \perfdown{0.66}{5.71} & \perfdown{19.73}{6.59} & 0.90 & \perfdown{0.012}{71.43}     \\ \midrule
\multirow{4}{*}[-3.5ex]{\rotatebox{90}{\shortstack{Forecasting at \\ Unobserved Location}}}  & TA-ANP & \textbf{26.24} & \textbf{30.56} & \textbf{0.69 }& \textbf{18.93} & \textbf{0.94} & \textbf{0.007 }       \\  \cmidrule{2-8}
                                                                          & w/o TAMQM & \perfdown{28.85}{9.95} & \perfdown{33.39}{9.26} & \perfdown{0.61}{11.59} & \perfdown{21.46}{13.37} & 0.85 & \perfdown{0.011}{57.14}        \\
                                                                          & w/o MC Dropout& \perfdown{30.26}{15.32} & \perfdown{35.31}{15.54} & \perfdown{0.58}{15.94} & \perfdown{23.13}{22.19} & 0.78 & \perfdown{0.019}{171.43}        \\
                                                                          & MC Dropout $\rightarrow$ Dropout & \perfdown{27.79}{5.91} & \perfdown{32.16}{5.24} & \perfdown{0.66}{4.35} & \perfdown{20.06}{5.97} & 0.90 & \perfdown{0.011}{57.14}     \\ \midrule
\multirow{4}{*}[-3.5ex]{\rotatebox{90}{\shortstack{Forecasting at \\ Observed Location}}}     & TA-ANP  & \textbf{15.57} & \textbf{19.99} & \textbf{0.89} & \textbf{11.18} & \textbf{0.96 }& \textbf{0.007}       \\  \cmidrule{2-8}
                                                                          & w/o TAMQM & \perfdown{17.29}{11.05} & \perfdown{22.01}{10.11} & \perfdown{0.87}{2.25} & \perfdown{12.40}{10.91} & 0.93 &  \perfup{0.006}{14.29}       \\
                                                                          
                                                                           & w/o MC Dropout & \perfdown{15.76}{1.22} & \perfdown{20.24}{1.25} & \perfsame{0.89}{0.00} & \perfdown{11.85}{5.99} & 0.97 & \perfup{0.006}{14.29}        \\

                                                                          & MC Dropout $\rightarrow$ Dropout & \perfdown{21.52}{38.21} & \perfdown{25.19}{26.01} & \perfdown{0.79}{11.24} & \perfdown{15.60}{39.53} & 0.88 & \perfdown{0.008}{14.29}     \\ 
                                                                          \bottomrule
\end{tabularx}
\end{table}

\subsection{Ablation study (Q6)}\label{sec:q6}
To systematically assess the contributions of the Task-Aware Multi-Query Mechanism (TAMQM) and Monte Carlo (MC) Dropout in TA-ANP, we conduct three controlled ablations: (1) removing TAMQM (w/o TAMQM); (2) removing MC Dropout (w/o MC Dropout); and (3) replacing MC Dropout with standard dropout (MC Dropout$\rightarrow$Dropout). As reported in \cref{tab:ablation}, both TAMQM and MC Dropout are critical components of TA-ANP. We analyze their effects below.

\begin{figure}[]
    \centering
    \includegraphics[width=0.8\textwidth]{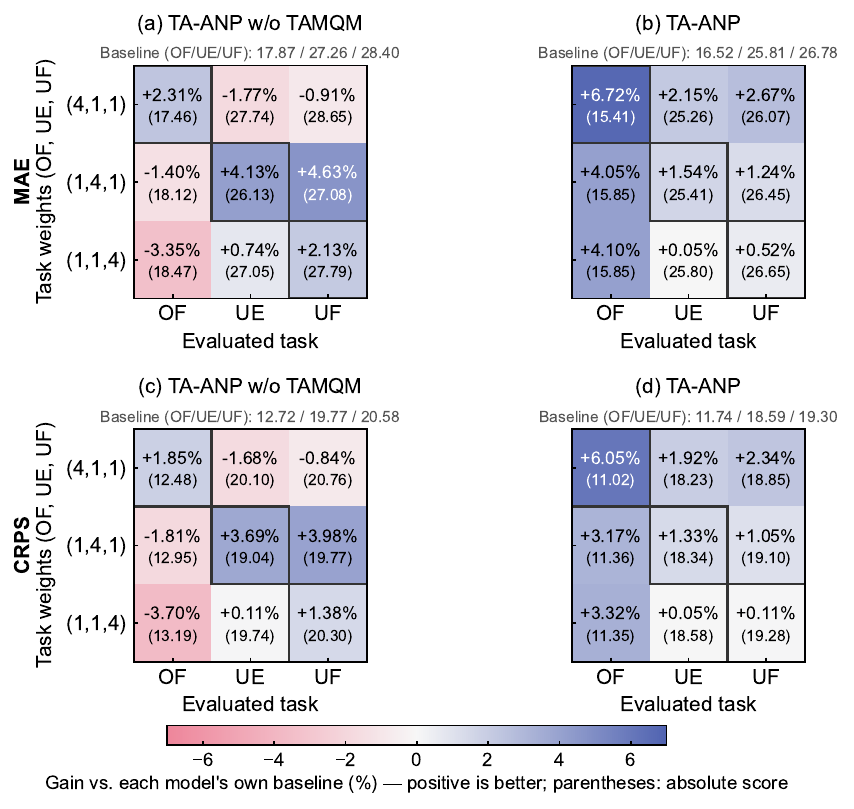}
    \caption{Task-weight perturbation analysis of cross-task interference. Each row emphasizes one of the three GTSI subtasks: observed-location forecasting (OF), unobserved-location estimation (UE), and unobserved-location forecasting (UF), relative to the task-balanced $(1,1,1)$ setting. Values report the performance gain relative to each model's own task-balanced baseline, where positive values indicate improvement and absolute scores are shown in parentheses.}
    \label{fig:task_weight_response}
\end{figure}
\subsubsection{Effect of the TAMQM}
TAMQM is designed to better accommodate the heterogeneous spatiotemporal dependencies of the three GTSI subtasks within a unified framework. Removing TAMQM leads to substantial degradation across all tasks. For example, on the unobserved road-segment inference task, MAE increases by 8.93\%, SMAPE increases by 8.26\%, and $R^2$ decreases by 10.00\%. TAMQM also benefits probabilistic inference, as removing it leads to marked deterioration in uncertainty-related metrics. These results demonstrate the overall effectiveness of task-specific query projections for joint GTSI modeling.

The above ablation confirms that TAMQM improves the joint modeling of all three GTSI subtasks. However, it does not directly reveal whether TAMQM alleviates the seesaw effect among them. To examine this more explicitly, we further conduct a task-weight perturbation experiment to introduce controllable seesaw effects across the three subtasks. We assign separate loss weights to observed-location forecasting (OF), unobserved-location estimation (UE), and unobserved-location forecasting (UF), denoted by $(\lambda_{\mathrm{OF}},\lambda_{\mathrm{UE}},\lambda_{\mathrm{UF}})$. Starting from the task-balanced setting $(1,1,1)$, we deliberately emphasize each subtask in turn using $(4,1,1)$, $(1,4,1)$, and $(1,1,4)$. This perturbation shifts the optimization preference toward one task, making potential cross-task performance trade-offs more directly observable. The same perturbations are applied to TA-ANP and its shared-query variant (TA-ANP w/o TAMQM) to examine whether TAMQM reduces the resulting interference.

As shown in \cref{fig:task_weight_response}, the shared-query variant exhibits clear cross-task interference. Emphasizing any subtask improves the emphasized task but degrades at least one of the other tasks. In terms of MAE, emphasizing OF yields a 2.31\% performance gain for OF while reducing the performance of UE and UF by 1.77\% and 0.91\%, respectively. Similarly, emphasizing UF improves UF performance by 2.13\% but reduces OF performance by 3.35\%. CRPS exhibits the same pattern, directly revealing cross-task interference and the seesaw effect. In contrast, TA-ANP largely maintains or improves the non-emphasized tasks under the same perturbations. 

Overall, the results show that TAMQM substantially alleviates the seesaw effect by reducing negative transfer among the three GTSI subtasks.

\subsubsection{Effect of MC Dropout}
As the key mechanism for uncertainty modeling, MC Dropout has the greatest impact on the unobserved road-segment inference task. For probabilistic prediction, removing MC Dropout increases CRPS by 22.69\% on unobserved segments, and the prediction interval coverage probability (PICP) drops from 0.94 to 0.78, far below the nominal 0.95. This suggests that MC Dropout effectively captures epistemic uncertainty via repeated stochastic forward passes, improving the quality of uncertainty estimates. In contrast, disabling MC Dropout has a smaller effect on segments with rich historical observations, where epistemic uncertainty is inherently lower.

Replacing MC Dropout with standard dropout also reduces performance, and we observe a clear task-dependent behavior. Compared with w/o MC Dropout (i.e., no dropout), standard dropout (active only during training) improves inference on unobserved segments but can slightly degrade prediction accuracy on observed segments. This difference is intuitive: unobserved-segment inference involves spatial extrapolation and relies heavily on generalization, where dropout regularization helps prevent overfitting; observed-segment prediction, however, is supported by sufficient historical data and favors stronger data fitting capacity, which can be weakened by dropout-induced randomness. These findings motivate future work on adaptive or task-specific dropout strategies.

\begin{figure}[]
    \centering
    \includegraphics[width=0.9\textwidth]{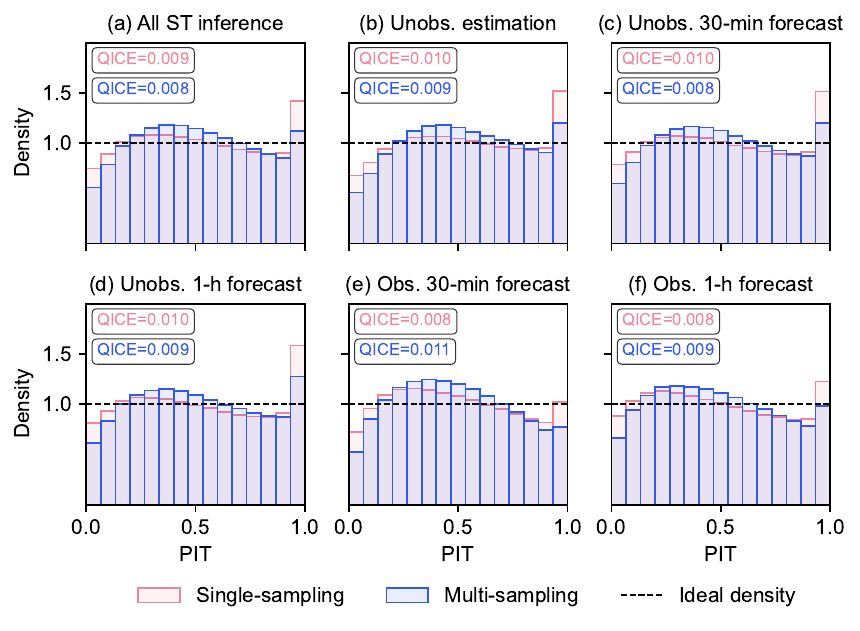}
    \caption{Probability integral transform (PIT) histograms comparing uncertainty calibration obtained with single-sampling (one stochastic forward pass; red) and multi-sampling (30 Monte Carlo dropout samples; blue) across six settings: (a) All spatiotemporal (ST) points inference. (b) Unobserved (Unobs.) estimation. (c) Unobs. 30-min forecast. (d) Unobs. 1-h forecast. (e) Observed (Obs.) 30-min forecast. (f) Obs. 1-h forecast. The dashed line indicates the ideal uniform density; QICE is reported in each panel (lower is better).}
    \label{fig:calibrationcomparsion}
\end{figure}

To further illustrate the effect of MC Dropout on uncertainty calibration, we evaluated probability integral transform (PIT) histograms. A PIT value is computed by inserting the ground-truth observation into the predicted cumulative distribution function. A perfectly calibrated probabilistic model yields PIT values that are uniformly distributed in $[0,1]$. As shown in \cref{fig:calibrationcomparsion}, with a single sample, PIT values cluster at high quantiles, indicating overconfidence with bias.\footnote{When the predictive distribution is overly concentrated (overconfident), realizations from the true data-generating process fall into the predictive tails more frequently than expected, producing excess PIT mass near 0 and 1. If the predictive distribution is also biased (e.g., shifted left relative to the truth), PIT values become systematically large, leading to a right-skewed PIT histogram.} Increasing the number of stochastic forward passes randomizes the model parameters during inference and produces diverse predictions in regions with high epistemic uncertainty, which broadens the predictive distribution and mitigates the overconfidence induced by single-sample inference. While 30 samples exhibit a slight underconfidence tendency, this is often preferable in downstream decision-making, as it provides more conservative and safer uncertainty bounds. By sweeping the number of samples from 1 to 50, we find that using 10 samples offers a good trade-off between computational cost (0.03 seconds per sample), predictive accuracy, and calibration in our setting.

Overall, the ablation results confirm the effectiveness and complementarity of the two core components in TA-ANP: TAMQM consistently improves all tasks by coordinating task-specific reliance on frequency-domain features, whereas MC Dropout is particularly important in high-uncertainty scenarios such as unobserved road segments. Their combination enables accurate and reliable global traffic status inference in complex traffic environments.

\section{Conclusion} \label{sec:con}
This paper tackles three interrelated challenges in global traffic state inference (GTSI). These challenges include resilience to multifaceted sensing disturbances, uncertainty quantification for trustworthy inference, and joint modeling of the three GTSI subtasks within a unified framework. To this end, we propose TA-ANP, which fuses floating car data with sparse fixed detectors, casts GTSI as a stochastic process, and introduces task-aware attention with differentiated spatiotemporal inductive biases.

Experiments on the metropolis-scale MMTD dataset lead to four main findings. First, across five independently generated sensor partitions, TA-ANP achieves best performance on all three GTSI subtasks under both deterministic and probabilistic metrics, demonstrating consistent performance across different sensing configurations. The gains are especially pronounced for uncertainty quantification at unobserved locations, where TA-ANP reduces CRPS and QICE by up to 23.47\% and 39.31\%, respectively, relative to the best baseline. Second, uncertainty quantified by TA-ANP is well calibrated against realized inference errors and provides actionable guidance for fixed-sensor placement in multi-source sensing settings. In particular, it requires about 20\% fewer sensors to reach comparable inference accuracy compared with graph-centrality heuristics. Third, TA-ANP remains robust under a compounded Damage--Repair--Addition lifecycle, achieving a performance retention ratio of 0.97 at peak disturbance compared with 0.90 for the strongest baselines, and it continues to benefit from incremental sensor deployments. Fourth, ablation results show that both TAMQM and MC Dropout are necessary. TAMQM mitigates the seesaw effect across subtasks, whereas MC Dropout is crucial for capturing epistemic uncertainty in high-uncertainty cases such as unobserved road segments.

Several limitations suggest promising directions for future work. First, the evaluation is performed on a single city; cross-city generalization therefore remains untested due to limited data, despite the neural process framework's theoretical ability to facilitate transfer. Second, the metrics used here capture numerical prediction error but do not directly measure downstream benefits for traffic control and management. Third, the resilience demonstrated in this study is largely passive. The model adapts to disturbances as they arise but does not actively leverage them. Looking ahead, future research could incorporate Taleb's concept of ``antifragility''~\citep{taleb2012antifragile} to move beyond passive resilience toward architectures that enable continual learning and improve through stochasticity and real-world perturbations~\citep{axenie2024Antifragility, sun2026Fragile}, ultimately supporting self-improving cyber-physical systems in the ITS context.

\printcredits


\section*{Acknowledgements}
This work was supported in part by the National Key R\&D Program of China (2023YFB4302600), the National Natural Science Foundation of China (52131202, 72350710798, 52272315, 62502406), the Provincial Key R\&D Program of Zhejiang (2024C01180), State Key Laboratory (SKL) of Biobased Transportation Fuel Technology, ZJU-YST joint research center for fundamental science, Guangdong and Hong Kong Universities ‘1+1+1’ Joint Research Collaboration Scheme, and Zhejiang University Global Partnership Fund. 

\appendix




\section{Definition of evaluation metrics}
\label{sec:metrics-define}
This appendix summarizes the evaluation metrics used in this study for completeness and reproducibility. All metrics are computed by excluding entries that are originally missing in the dataset. Let $\{(y_i,\hat y_i)\}_{i=1}^{M}$ denote the ground-truth and point prediction pairs over the $M$ valid (non-missing) samples. 

\begin{itemize}
    \item \textbf{Relative Root Mean Square Error (RRMSE):}
    \begin{equation}
        \mathrm{RRMSE}  = \frac{\sqrt{\frac{1}{M} \sum_{i=1}^{M} (y_i - \hat{y}_i)^2}}{\frac{1}{M} \sum_{i=1}^{M} y_i}
    \end{equation}

    \item \textbf{Symmetric Mean Absolute Percentage Error (SMAPE):}
    \begin{equation}
        \mathrm{SMAPE} = \frac{100\%}{M} \sum_{i=1}^{M} \frac{|y_i - \hat{y}_i|}{(|y_i| + |\hat{y}_i|) / 2}
    \end{equation}

    \item \textbf{Continuous Ranked Probability Score (CRPS)} \citep{matheson1976CRPS}:
    CRPS simultaneously assesses the accuracy (deviation from observations) and sharpness (concentration of the distribution) of the predictive distribution, and is widely used for evaluating probabilistic forecasts. A lower CRPS indicates higher quality of the predictive distribution. It is defined as:
    \begin{equation}
        \mathrm{CRPS}(F, y) = \int_{-\infty}^{\infty} (F(x) - \mathbb{I}(x \ge y))^2 dx
    \end{equation}
    where \(F\) is the Cumulative Distribution Function (CDF) predicted by the model, \(y\) is the true observed value, and \(\mathbb{I}(\cdot)\) is the indicator function.
  
    \item \textbf{Prediction Interval Coverage Probability (PICP)}: This metric evaluates the coverage of prediction intervals for the true values and is defined as:
    \begin{equation}
        \mathrm{PICP} = \frac{1}{M} \sum_{i=1}^{M} \mathbb{I}(q_{\alpha/2} \le y_i \le q_{1-\alpha/2})
    \end{equation}
    where  \(\alpha\) is the significance level, and \(q_{\alpha/2}\) and \(q_{1-\alpha/2}\) represent the \(\alpha/2\) and \((1-\alpha/2)\) quantiles of the predictive distribution at the \((1-\alpha)\) confidence level, respectively. Ideally, PICP should be close to the confidence level \((1-\alpha)\) of the prediction interval. In this study, we set \(\alpha=0.05\).

    \item \textbf{Mean Prediction Interval Width (MPIW)}:
    MPIW measures the average width of the prediction intervals and characterizes their sharpness. For a prediction interval with nominal coverage level $(1-\alpha)$, it is defined as
    \begin{equation}
        \mathrm{MPIW}
        =
        \frac{1}{M}
        \sum_{i=1}^{M}
        \left(
        q_{1-\alpha/2}^{(i)} - q_{\alpha/2}^{(i)}
        \right),
    \end{equation}
    where $q_{\alpha/2}^{(i)}$ and $q_{1-\alpha/2}^{(i)}$ denote the lower and upper quantiles of the predictive distribution for sample $i$, respectively. In this study, we set $\alpha=0.05$, consistent with the PICP evaluation, and therefore compute MPIW over the 95\% prediction intervals. 
        
    \item \textbf{Quantile Interval Calibration Error (QICE)} \citep{Han2022QICE}: Compared to PICP, QICE comprehensively assesses the calibration between the predictive distribution and the true distribution by integrating the coverage deviations across all quantile intervals. It is defined as:
    \begin{equation}
        \mathrm{QICE} = \frac{1}{N} \sum_{n=1}^N \left| r_n - \frac{1}{N} \right|,
    \end{equation}
    where \(N\) is the number of quantile intervals, and \(r_n\) denotes the actual coverage of the \(n\)-th quantile interval, calculated as:
    \begin{equation}
        r_n = \frac{1}{M_n} \sum_{i=1}^{M_n} \mathbb{I}(y_i \geq \hat{y}_{\mathrm{low}}^n) \cdot \mathbb{I}(y_i \leq \hat{y}_{\mathrm{high}}^n),
    \end{equation}
    where \(M_n\) is the number of samples in the \(n\)-th partition, \(y_i\) is the observed value of the \(i\)-th sample, and \(\hat{y}_{\mathrm{low}}^n\) and \(\hat{y}_{\mathrm{high}}^n\) are the lower and upper boundaries corresponding to the \(n\)-th quantile interval, respectively. Ideally, the QICE value should be close to 0, indicating that the actual coverage of different quantile intervals is perfectly consistent with the theoretical target coverage.
\end{itemize}

\section{Architectural hyperparameter settings for TA-ANP} 
\label{sec:A2}
\cref{tab:a2} summarizes the key architectural hyperparameters of TA-ANP, organized by the components introduced in Section 4.

\begin{table}[t]
\centering
\caption{Architectural hyperparameter settings for TA-ANP.}
\label{tab:a2}
\begin{tabular}{lll}
\toprule
Component (Section) & Hyperparameter & Value \\
\midrule
Encoder (Sec.~4.1) & Representation dimension ($\mathbf{r}_i$) & 64 \\
Latent path (Sec.~4.1) & Latent variable dimension ($\mathbf{z}$) & 64 \\
Cross-attention (Sec.~4.2) & Number of attention heads & 4 \\
Cross-attention (Sec.~4.2) & Key/value projection dimension & 64 \\
TAMQM (Sec.~4.2) & Task-specific query projection dimension & 64 \\
MC Dropout (Sec.~4.3) & Dropout rate ($p$) & 0.1 (encoder) / 0.2 (decoder) \\
MC Dropout (Sec.~4.3) & Number of stochastic forward passes ($K$) & 30 \\
\bottomrule
\end{tabular}
\end{table}

\section{Baseline adaptation details}
To ensure a fair comparison under the GTSI setting considered in this work, while preserving the core design principles of each baseline, we made the following adjustments. First, the output layer of each baseline was adjusted to jointly produce state estimation at unobserved locations and forecasting at both observed and unobserved locations, consistent with the three GTSI subtasks defined in this work; for AGCRN and DeepSTUQ, both originally designed for fixed-node point forecasting, we additionally adopted the inductive training strategy of~\citet{zhou2025MoGERNNInductiveTraffic} to enable generalization to unobserved locations, while for MoGERNN, whose original autoregressive decoder does not natively support estimation at unobserved locations, we replaced it with a multilayer perceptron output head. Second, the graph adjacency used by STCAGCN, DGAE, and MoGERNN was reconstructed using the road-segment identity vectors defined in Section 3. Third, the input feature layer of every baseline was extended to incorporate the multi-source features (FCD flow, FCD speed, and static road attributes) used in this work, with STCAGCN additionally incorporating FCD flow alongside the speed-derived features it originally relies on. For DistPred, the learning objective of its IGNNK backbone was further adapted to directly predict future states at all nodes rather than performing kriging interpolation alone. The brief introduction of adopted baselines are as follows.
\begin{itemize}

\item  \textbf{AGCRN} \citep{LeiAGCRN20}: An adaptive graph convolutional recurrent network for fixed-node traffic prediction, which learns node-adaptive parameters to capture dynamic spatial dependencies and employs a GRU for temporal modeling.

\item  \textbf{STCAGCN} \citep{nie2023flowestimation}: A traffic-volume estimation model that constructs a speed-informed adaptive graph adjacency matrix to capture spatial dependencies among road segments.

\item  \textbf{DGAE} \citep{zhou2025NetworkWidea}: An inductive graph auto-encoder for traffic state estimation at unobserved locations, which generates embeddings for unobserved nodes via Dirichlet Energy-based Feature Propagation for Directed graphs (DEFP4D) applied in the auto-encoder's latent space.

\item  \textbf{MoGERNN} \citep{zhou2025MoGERNNInductiveTraffic}: An inductive spatio-temporal graph model that forecasts future traffic states at both observed and unobserved locations, combining a Mixture of Graph Experts (MoGE) module with a GRU-based encoder--decoder.

\item  \textbf{UIGNN} \citep{Mei2023UIGNN}: An uncertainty-aware global traffic prediction model that combines evidential regression with an inductive graph kriging network.

\item  \textbf{DeepSTUQ} \citep{qian2023UncertaintyQuantificationTraffic}: An uncertainty-aware, fixed-node traffic forecasting framework.

\item  \textbf{DistPred} \citep{Liang2025}: A plug-and-play, distribution-free uncertainty quantification framework applied on top of a backbone predictor; in this work, we pair it with IGNNK \citep{wu2021IGNNK}, an inductive graph neural network originally designed for spatiotemporal kriging.

\end{itemize}

\begin{figure}[]
    \centering
    \includegraphics[width=\textwidth]
    {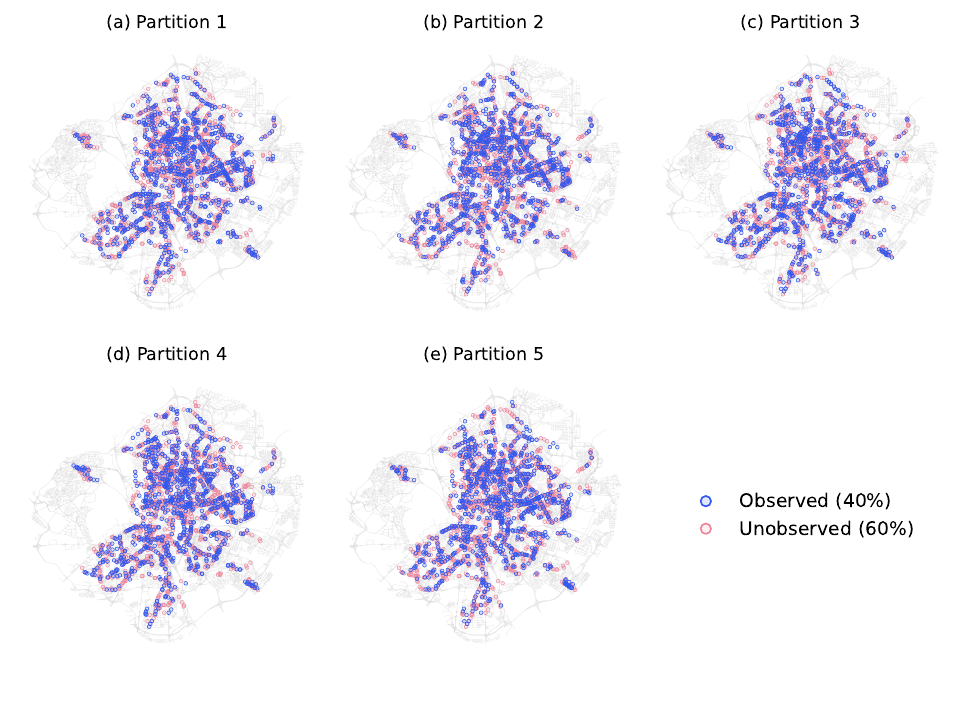}
    \caption{Spatial distributions of the observed and unobserved locations
obtained from five independent repetitions of the uniform
observed-location sampling procedure. Blue circles denote observed
locations, whereas red circles denote unobserved locations. Each
configuration contains 949 observed and 1,422 unobserved locations. Partitions 1--5 correspond to random seeds 42, 52, 62, 72, and 82.}
    \label{fig:sensor_partitions_five}
\end{figure}

\section{Spatial Distributions of Sensor Partitions}
\label{app:sensor_partitions}

\cref{fig:sensor_partitions_five} presents the spatial distributions of the observed and unobserved locations in five independently generated observation configurations. In each configuration, 40\% of the 2,371 road segments equipped with a fixed sensor (949 locations) are uniformly sampled as observed locations, while the remaining 60\% (1,422 locations) are treated as unobserved locations by excluding their fixed-sensor measurements from the model inputs. Across all ten pairwise comparisons, the proportion of locations
with different observed/unobserved statuses ranges from 47.1\% to 49.3\%.

\bibliographystyle{cas-model2-names}

\bibliography{cas-refs.bib}

\end{document}